%% file: main.tex
\documentclass[10pt,twocolumn]{article}

\usepackage[margin=0.72in]{geometry}
\usepackage[T1]{fontenc}
\usepackage{lmodern}
\usepackage{microtype}
\usepackage{amsmath,amssymb,amsthm}
\usepackage{booktabs}
\usepackage{graphicx}
\usepackage[font=small,labelfont=bf,hypcap=false]{caption}
\usepackage[numbers,sort&compress]{natbib}
\usepackage{xurl}
\graphicspath{{figures/}{figures/generated/}}

\title{Boundary Density Likelihood for Direct Event-Time Supervision}
\ifdefined\ANONYMOUSPAPER
\author{Anonymous Authors}
\else
\author{Clark Peng \and Tolga Din\c{c}er}
\fi
\date{}

\newcommand{\xseq}{x_{1:T}}
\newcommand{\rate}{\lambda}
\newcommand{\kernel}{\kappa}
\newcommand{\Rnonneg}{\mathbb{R}_{\ge 0}}
\newcommand{\methodteaser}{\includegraphics[width=\textwidth]{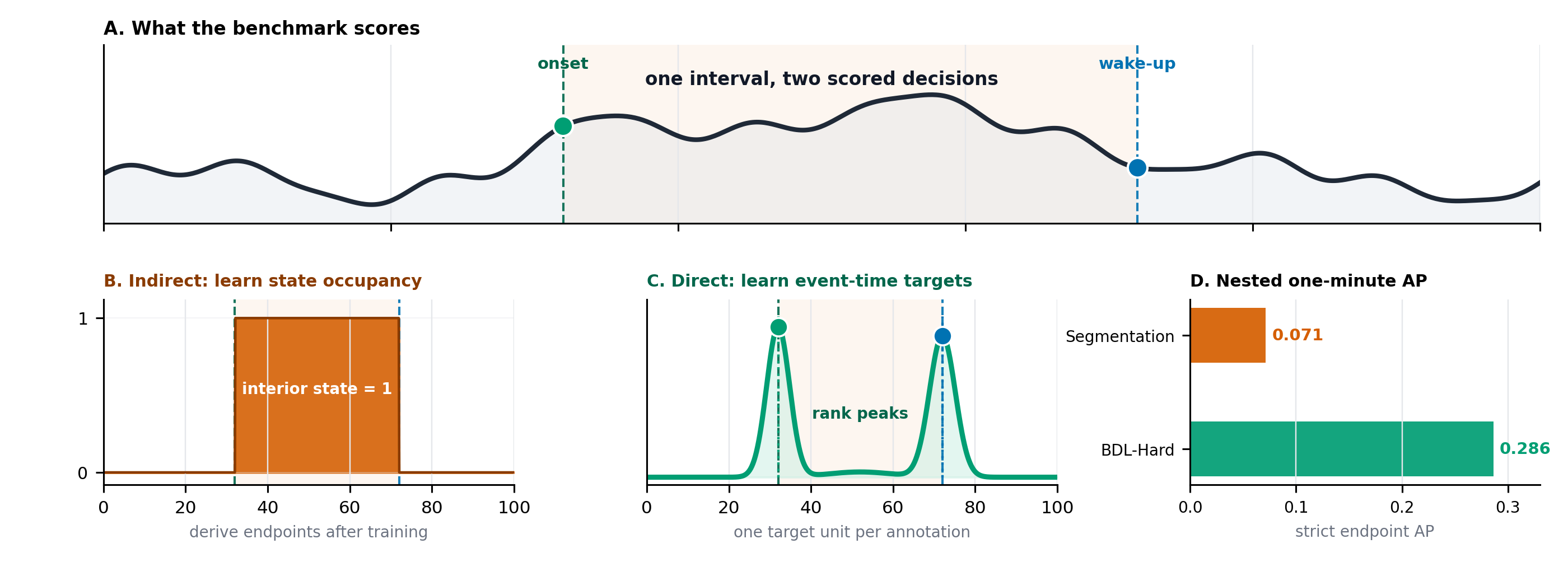}}

\begin{document}
\twocolumn[
\begin{@twocolumnfalse}
\maketitle

\begin{abstract}
\input{sections/abstract}
\end{abstract}
\vspace{0.55em}
\begin{center}
\methodteaser
\captionof{figure}{Direct event-time prediction matches the benchmark output.
One sleep interval produces two scored events: segmentation learns occupancy and derives endpoints afterward, whereas BDL ranks endpoint scores directly.
Across five nested folds, BDL-Hard improves strict one-minute AP from 0.071 to 0.286 ($4.0\times$) and mAP from 0.586 to 0.705 (+11.9 percentage points).}
\label{fig:method-overview}
\end{center}
\vspace{0.4em}
\end{@twocolumnfalse}
]

\input{sections/introduction}
\input{sections/method}
\input{sections/related_work}
\input{sections/experiments}
\input{sections/results}
\input{sections/limitations}
\input{sections/conclusion}

\bibliographystyle{plainnat}
\begingroup
\footnotesize
\sloppy
\renewcommand{\baselinestretch}{0.76}\selectfont
\providecommand{\doi}[1]{doi: \url{#1}}
\renewcommand{\doi}[1]{doi: \url{#1}}
\setlength{\bibhang}{0.9em}
\setlength{\bibsep}{0pt plus 0.2ex}
\bibliography{main}
\endgroup

\clearpage
\onecolumn
\appendix
\input{sections/appendix}

\end{document}

%% file: sections/abstract.tex
Event detection turns long recordings into a sparse set of ranked timestamps.
Yet many sequence models are trained for samplewise segmentation and only convert predicted states into events after training.
We ask whether training directly for the evaluated output improves detection.
Boundary Density Likelihood (BDL) assigns one unit of target mass to each annotated event, preserves that mass through smoothing and temporal downsampling, and uses a Poisson objective to estimate expected event mass in each output bin; local peaks become ranked detections.
In a prespecified five-fold nested sleep study, BDL-Hard raises pooled out-of-fold mAP from 0.586 to 0.705 over interval segmentation (+11.9 percentage points; 95\% interval $[10.8,13.0]$) and strict one-minute AP from 0.071 to 0.286 ($4.0\times$; +21.5 points), improving on every outer fold.
A separate held-out rerun reproduces the direction of the effect.
A matched boundary-BCE detector reaches 0.702 mAP, showing that direct boundary supervision and event decoding account for most of the gain, with a smaller contribution from the Poisson objective.
The same trend appears with an offline convolutional model; causal, Transformer, and patient-grouped seizure experiments leave broader generalization unresolved.
Taken together, the results support training on event times when timestamps, rather than interval occupancy, define the evaluated output.

%% file: sections/introduction.tex
\section{Introduction}
\label{sec:introduction}

Event detection compresses a long recording into a small set of ranked decisions.
A detector can label nearly every sample correctly and still fail if an event is early, late, duplicated, missed, or poorly ranked.
This distinction matters in clinical monitoring, activity analysis, industrial logs, audio, and scientific measurement, where timed events trigger alerts, delimit episodes, or select segments for review \citep{guralnik1999,aminikhanghahi2017,shoeb2010,goldberger2000,alexander2017,esper2023,azib2023universal}.
The corresponding benchmarks sort detections by confidence, match them to annotations inside tolerance windows, and summarize precision and recall.
The central modeling question is therefore what each dense output is trained to represent.

The common surrogate is samplewise segmentation.
Interval labels become dense states, a sequence model predicts those states, and event scores are recovered afterward from thresholds or transitions.
That reduction is appropriate when deployment consumes state occupancy.
It is indirect when evaluation consumes ranked onset, offset, or alarm times.
Sleep detection makes the mismatch concrete: an interval may last for hours, while event AP gives credit only for onset and wake-up predictions inside minute-scale windows.
The interior supplies context, but the benchmark scores the endpoints.

We formulate \emph{Boundary Density Likelihood} (BDL), a direct event-time approach that retains the same dense sequence-model interface.
Each annotation defines one target unit.
A unit-sum timing kernel can redistribute that unit, temporal binning sums it onto the model timeline, and a Poisson mean score fits the conditional mean per-bin target: a count for hard targets and kernel-assigned target mass after smoothing.
Local maxima become ranked event detections.
The construction applies to empty windows, repeated events, point events, onset-only labels, interval endpoints, and multi-type streams without requiring paired endpoints.

We make three contributions.
First, BDL gives localized event targets an additive unit that remains interpretable through smoothing and temporal downsampling, then fits the resulting per-bin conditional mean with a Poisson score.
Second, a frozen five-fold nested sleep study shows that the boundary detector raises strict one-minute AP to $4.0\times$ the segmentation baseline and improves mAP by 11.9 percentage points, with a positive difference on every outer fold; a separate holdout and convolutional U-Net development studies show the same direction.
This detector comparison changes the target, output representation, and decoder together; it asks which formulation better serves event-time evaluation rather than assigning the entire difference to one component.
Third, matched boundary-BCE controls retain 97\% of the nested mAP gain and decompose that result: the incremental Poisson--Bernoulli difference is smaller, target-dependent, and not stable in the patient-grouped CHB-MIT study.

Localized supervision itself has broad precedent: pose, audio, and temporal-action systems emphasize keypoints, sound events, starts, or ends, and recent time-series work places regression targets near annotated events \citep{newell2016stacked,xiao2018simple,luo2021,yu2022,phan2015,gao2017turn,lin2018bsn,lin2019bmn,azib2023universal}.
Our contribution is not the first localized map.
It is the combination of an additive temporal unit, sum-preserving discretization, a conditional-mean score, and controlled evidence separating the boundary formulation from its scoring rule.

%% file: sections/method.tex
\section{Boundary Density Likelihood}
\label{sec:method}

BDL keeps the usual sequence-model interface---an input window maps to per-bin predictions---but changes what those predictions represent.
For bin $b$ and event type $c$, the output estimates the conditional mean of the constructed target: a count for hard targets and kernel-assigned event mass for smoothed targets.
We use ``boundary'' to mean the scored event time: for interval data this is an onset or offset, and for point-event data it is the event itself.
Here ``density'' denotes a nonnegative quantity that adds over time in the target's event unit; outputs are not normalized as a probability distribution over one unknown position in the window.
Figure~\ref{fig:target-pipeline} summarizes the path from annotations to the Poisson objective on model output bins.

\begin{center}
\includegraphics[width=\columnwidth]{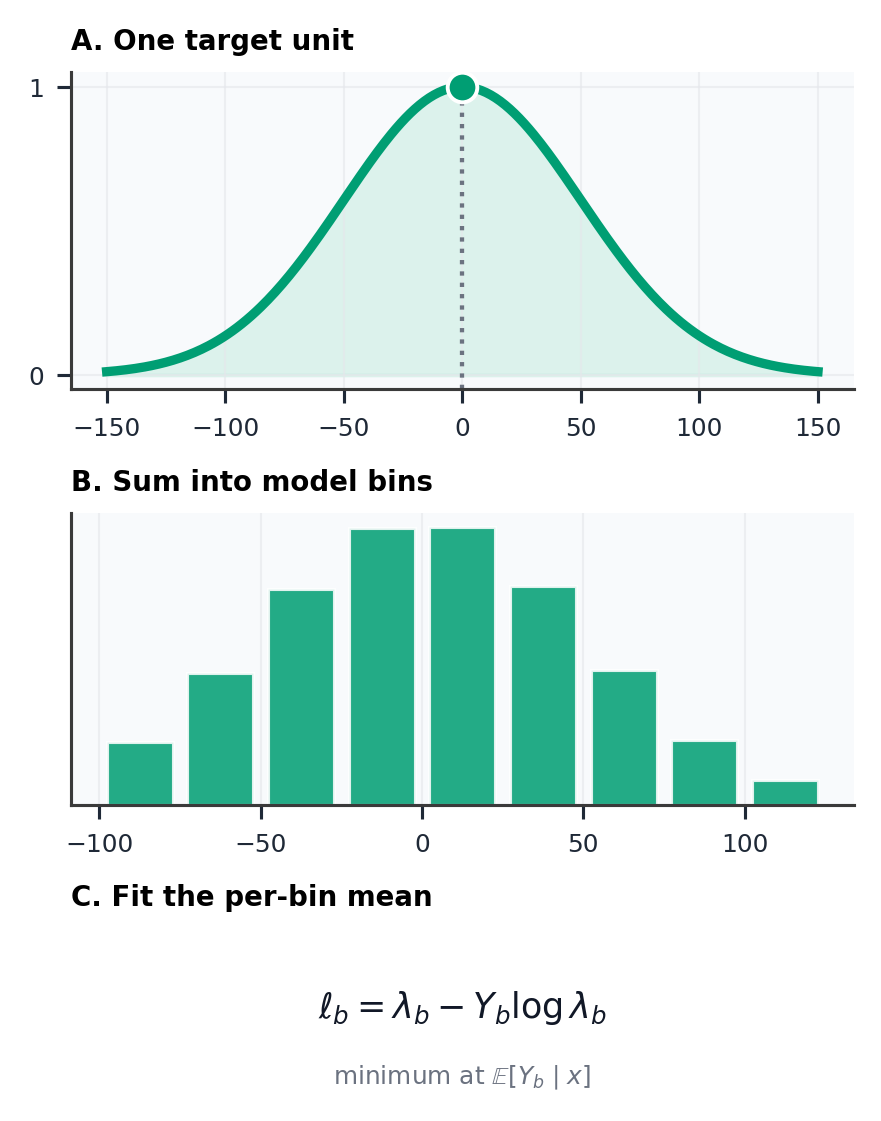}
\captionof{figure}{Each annotation contributes one target unit; smoothing redistributes it, binning sums it, and the Poisson score fits its per-bin conditional mean.}
\label{fig:target-pipeline}
\end{center}

\subsection{Scored Temporal Detections}
\label{subsec:scored-detections}

Let $\xseq$ be a multivariate time series on a discrete timeline.
We use $c$ to index an event type and its corresponding output channel: a point event, an onset, an offset, or another application-specific event.
For event type $c\in\mathcal{C}$, let $B_c$ denote the annotated event times as a finite indexed collection (and hence a multiset) over $\{1,\ldots,T\}$.
Point-event, onset-only, and multi-type datasets use the event types they need.
An interval event of type $q$ with endpoints $(a_{j,q},b_{j,q})$ can be represented as two scored event types,
\begin{align}
    B_{q,\mathrm{on}} = \{a_{j,q}\}_{j=1}^{n_q},
    \qquad
    B_{q,\mathrm{off}} = \{b_{j,q}\}_{j=1}^{n_q}.
\end{align}
The likelihood therefore does not require paired endpoints.
If an application needs valid on--off intervals, alternation or pairing is a decoder constraint applied after the event times have been scored.

A detector emits scored temporal detections
\begin{align}
    \widehat{B}_c = \{(\hat{\tau}_{k,c}, s_{k,c})\}_{k=1}^{m_c},
\end{align}
where $\hat{\tau}_{k,c}$ is a proposed event time and $s_{k,c}$ is the ranking score.
Event-level average precision sorts these predictions, matches them to nearby unmatched annotations, and counts a true positive only when the prediction falls inside a task-specific tolerance window \citep{everingham2010pascal,lin2014coco,salles2023,tatbul2018precision}.
The matched object is an event occurrence in time: a window may contain zero, one, or many such occurrences, and the metric consumes sparse scored detections rather than per-sample labels.
The corresponding training signal assigns hard counts or smoothed event mass
to each output bin.

\subsection{Event Occurrences on the Output Timeline}
\label{subsec:targets}

For event type $c$, the annotations define a counting measure over time,
\begin{align}
    \mu_c = \sum_{\tau\in B_c}\delta_{\tau},
\end{align}
where $\delta_\tau$ denotes one count placed at event time $\tau$.
This is the standard counting-process view of event data: a time region contains the number of event occurrences that happened there \citep{kingman1993,shchur2021neural}.
BDL uses this counting measure for supervision without requiring a full generative point-process model.

Annotation times may be uncertain, discretized, or only meaningful up to the tolerance of the benchmark.
BDL can express that uncertainty by replacing each point mass with a nonnegative discrete kernel $\kernel:\mathbb{Z}\rightarrow\Rnonneg$ whose values sum to one,
\begin{align}
    \sum_{r\in\mathbb{Z}}\kernel(r)=1.
\end{align}
The smoothed target on the original timeline is
\begin{align}
    y_c(t) = \sum_{\tau\in B_c}\kernel(t-\tau),
    \qquad t=1,\ldots,T.
    \label{eq:smoothed-target}
\end{align}
If the finite target timeline clips the kernel support at a recording boundary, the visible support is renormalized so that the annotation continues to contribute one event.
Smoothing changes where the annotation's unit contribution is assigned, not
its total contribution on the target timeline.
The resulting value can be zero, fractional after smoothing, one for a hard event, or larger than one when multiple events occupy the same region.

We use three kernel families:
\begin{align}
    \kernel_{\mathrm{hard}}(r) &= \mathbf{1}\{r=0\}, \\
    \kernel_{\mathrm{gau}}(r) &\propto
        \exp\left(-\frac{r^2}{2\sigma^2}\right)\mathbf{1}\{|r|\le 3\sigma\}, \\
    \kernel_{\mathrm{tol}}(r) &\propto
        \frac{1}{K}\sum_{k=1}^{K}\mathbf{1}\{|r|\le \delta_k\}.
\end{align}
The tolerance-aligned kernel uses the same tolerance set $\{\delta_k\}_{k=1}^{K}$ as the event-localization metric before normalizing the kernel sum.
It provides a benchmark-scale label-uncertainty model rather than a differentiable AP objective.
Because every kernel sums to one, $\sum_t y_c(t)=|B_c|$ after finite-timeline renormalization.

The same interpretation determines how supervision moves to a coarser output timeline.
If the sets $G_b$ form a disjoint partition of the original timeline into model output bins, BDL defines
\begin{align}
    Y_c(b) = \sum_{t\in G_b} y_c(t).
    \label{eq:downsampled-target}
\end{align}
Equation~\ref{eq:downsampled-target} integrates over the bin: $Y_c(b)$ is an
event count for a hard target and may be fractional after smoothing, while
summing over bins recovers the number of visible annotations.
Summing into bins gives BDL a stable unit, unlike localized objectives whose numerical scale changes with kernel width, clipping, or pooling.
Output stride remains a detector choice: it should preserve the metric's localization resolution and keep same-type event collisions rare.
BDL can represent more than one occurrence in a bin, but a decoder cannot recover their individual times after they have been merged.
Appendix~\ref{app:derivation} derives the resolution and sparsity checks used for the reported strides.

\subsection{Poisson Mean Score and Decoding}
\label{subsec:poisson}

Let a sequence model with parameters $\theta$ output logits $z_{\theta,c}(b)$.
BDL converts these logits to nonnegative mean-target predictions,
\begin{align}
    \rate_{\theta,c}(b\mid\xseq)
    =
    \operatorname{softplus}\left(z_{\theta,c}(b) + \operatorname{softplus}^{-1}(\rate_0)\right)
    + \epsilon,
    \label{eq:rate-param}
\end{align}
where $\rate_0$ initializes the output near a sparse-count prior and $\epsilon>0$ prevents numerical singularities.
The reported runs set $\rate_0=\Delta/L$, where $\Delta$ is the output stride in raw samples and $L$ is a fixed benchmark reference spacing; for sleep, $L=17{,}280$ samples (one day), so $\rate_0=5.79\times10^{-4}$ at $\Delta=10$.

For one bin and event type, write $Y=Y_c(b)$ and $\rate=\rate_{\theta,c}(b\mid\xseq)$.
For an integer event count, the Poisson negative log-likelihood is
\begin{align}
    \ell_{\mathrm{P}}(Y,\rate)
    =
    \rate - Y\log \rate + \log \Gamma(Y+1).
\end{align}
For smoothed fractional targets, we use its parameter-dependent part as a
Poisson deviance, or quasi-likelihood, score.
The $\log\Gamma(Y+1)$ extension is target-only and does not affect optimization,
but it should not be interpreted as a Poisson probability mass function over
fractional observations.
With $\rate_{c,b}=\rate_{\theta,c}(b\mid\xseq)$, the empirical loss is
\begin{align}
    \mathcal{L}_{\mathrm{BDL}}(\theta)
    =
    \sum_c\sum_b
    \left[
        \rate_{c,b}
        -
        Y_c(b)\log \rate_{c,b}
    \right].
    \label{eq:bdl-loss}
\end{align}
Zero-event windows contribute only the prediction penalty, while repeated events contribute additional count.

The per-bin Poisson score for integer counts, and the corresponding deviance
score for fractional targets, give the target construction its statistical meaning.
Unit-sum kernels and sum binning ensure $\sum_b Y_c(b)=|B_c|$, and the conditional expected risk is minimized at $\rate^*(b,c\mid\xseq)=\mathbb{E}[Y_c(b)\mid\xseq]$.
This statement ignores the numerical floor in Equation~\ref{eq:rate-param}; with that floor, the optimum is clipped at the lower bound, with an infimum at $\epsilon$ when the conditional mean is no larger than $\epsilon$.
Summing the binwise terms is a composite Poisson score unless conditional independence across bins is assumed.
The fitted quantity is therefore neither a normalized position distribution
nor an unscaled proximity score; it is $\mathbb{E}[Y_c(b)\mid\xseq]$, where
$Y_c(b)$ is a count for a hard target and kernel-assigned event mass for a
smoothed target.
Appendix~\ref{app:derivation} gives the conditional-mean and deviance derivations.

At inference time, each learned score sequence is decoded by ranking local maxima.
Domain constraints, such as alternating onset and offset labels for interval datasets, are applied after the score has produced candidate times.
The objective estimates a per-bin target closer to the benchmark output than interval occupancy, but it is not proved consistent for tolerance-matched AP: peak finding, ranking, duplicate suppression, and one-to-one matching remain detector and evaluator operations.

%% file: sections/related_work.tex
\section{Related Work}
\label{sec:related-work}

\paragraph{Event-level evaluation.}
Event detection and changepoint detection have long been treated separately from samplewise classification because a long sequence may matter primarily through the events it contains \citep{guralnik1999,aminikhanghahi2017}.
Modern benchmarks increasingly use the ranked-matching structure of object and activity detection: predicted times are sorted by confidence, matched to annotations under a tolerance rule, and summarized by precision--recall or average precision \citep{everingham2010pascal,lin2014coco}.
SoftED and range-aware precision--recall make the same point for time series, where per-sample accuracy can disagree with event-level utility \citep{salles2023,tatbul2018precision}.
This paper carries that evaluation view into the training target.
If the metric ranks matched events, the model can supervise those events directly rather than first learning a state label at every sample.

\paragraph{Segmentation and localized supervision.}
Samplewise segmentation models remain practical because they share computation across long recordings, and segmentation architectures are well established in sleep and biomedical time series \citep{supratak2017,perslev2019,phan2019,ronneberger2015}.
The relevant choice is the loss attached to such sequence predictors when the benchmark consumes detections.
Pose estimation, acoustic-event detection, and temporal action localization provide useful precedents for placing supervision near annotated times through keypoint maps, proximity targets, or start/end evidence \citep{newell2016stacked,xiao2018simple,luo2021,yu2022,phan2015,gao2017turn,lin2018bsn,lin2019bmn}.
Density-map counting likewise turns point annotations into nonnegative maps whose integral estimates a count, and later work connects density quality to localization \citep{lempitsky2010count,wan2021generalized}.
Recent time-series event work makes a similar criticism of samplewise event/non-event labels, especially for rare and interval-defined events, and argues for a broad regression-based alternative to classification \citep{azib2023universal}.
BDL combines these ideas for temporal event AP while making the target integral explicit: every annotation contributes one additive unit, smoothing redistributes it, sum binning moves it to the model timeline, and a Poisson score fits the per-bin conditional mean.
The emphasis is not localized supervision alone, but a target unit with the same interpretation after edge normalization and temporal discretization, including windows with no events or multiple nearby events.

\paragraph{Counting-process likelihoods.}
Temporal point processes provide the standard language of counting measures, conditional intensities, and likelihoods over event times \citep{kingman1993,du2016rmtpp,mei2017neural,shchur2021neural}.
BDL borrows the counting-measure view for supervised prediction: given an observed window, it predicts the conditional mean hard count or smoothed target mass in each output bin without modeling censoring, future histories, survival terms, or mark distributions.
Without an output floor, the parameter-dependent Poisson score is minimized at this conditional mean and reduces to the per-bin Poisson negative log-likelihood for integer counts \citep{gneiting2007proper}.
This conditional-mean property does not imply optimal ranking for event AP.
This places BDL between segmentation and full event-history modeling: it keeps an efficient dense sequence interface while making each output the conditional mean of an additive event target under a Poisson or quasi-likelihood score.

%% file: sections/experiments.tex
\section{Experiments}
\label{sec:experiments}

The experiments follow the paper's two-level claim.
First, does a detector trained to predict boundaries outperform one trained to segment interval occupancy when both are evaluated as ranked events?
Second, after fixing the boundary target, output channels, and decoder, does Poisson scoring improve on BCE?
Sleep onset and wake-up detection is the central setting; held-out and nested studies test the two questions, while CHB-MIT, point-event, architecture, stride, streaming, and state-reconstruction analyses define scope.
Within each matched study, we fix the split, backbone, event evaluator, and decoder-search budget; Appendix~\ref{app:experimental-details} gives preprocessing, optimization, and calibration details.

The Child Mind Institute sleep benchmark contains 277 wrist-accelerometer recordings annotated with onset and wake-up events \citep{alexander2017,migueles2019,robbins2019,esper2023}.
Its intervals can last hours, while event AP scores endpoints within one-to-thirty-minute windows.
CHB-MIT contains 256-Hz EEG recordings with seizure intervals and second-scale endpoint tolerances \citep{shoeb2010,goldberger2000,klem1999}.
We use the 141 seizure-positive recordings, so this study evaluates endpoint localization conditional on a positive recording rather than screening false alarms on seizure-free EEG.
\IfFileExists{results/generated/point_event_ablation_ready.tex}{Appendix studies on Martian bow shock and credit-card fraud remove paired interval endpoints \citep{azib2023universal}.}{}

The four-fold sleep development study fixes a bidirectional GRU and compares BDL-Hard, BDL-Gaussian, and BDL-Tolerance with cross-entropy, class-weighted, and focal segmentation \citep{krawczyk2016,lin2017focal}.
It uses a shared learning rate chosen for stable Poisson training, so the results are matched configurations rather than objective-specific optima.
A held-out rerun gives BDL, boundary BCE, segmentation, and localized MSE the same learning-rate, checkpoint, and decoder-search budgets, selects each on calibration data, and scores the selected models on a separate test partition.
A frozen five-fold outer study adds disjoint inner calibration and three final seeds per fold; it compares BDL-Hard with interval segmentation on identical outer and inner splits and crosses hard and Gaussian boundary targets with Poisson and Bernoulli scores.
U-Nets, an attention-gated U-Net, a Transformer, and a forward GRU test architecture and streaming limits \citep{ronneberger2015,oktay2018attention,vaswani2017attention}.
CHB-MIT uses descriptive recording-split detector studies plus a separate patient-grouped Poisson--BCE analysis.

BDL-Hard places one target unit at the annotated time; Gaussian and tolerance variants redistribute that unit before sum binning.
Appendix~\ref{app:experimental-details} documents finite-recording and crop-edge handling; the headline BDL-Hard result is unaffected by either issue.

Candidate smoothing, peak selection, thresholds, and interval alternation are selected on validation or calibration data before outer-test scoring.
Sleep mAP averages onset and wake-up AP over one-to-thirty-minute tolerances; seizure mAP uses one-to-sixty-second tolerances, with one-to-three-second AP as the strict endpoint regime.
The main studies split by recording, while the grouped CHB-MIT study splits by patient.
Output stride is part of the detector; Appendix~\ref{app:derivation} reports its resolution and sparsity checks.

%% file: sections/results.tex
\section{Results}
\label{sec:results}

\input{sections/result_figures}

\subsection{Direct Boundary Detection Improves Strict Event AP}

The frozen five-fold nested comparison provides the central detector-level result.
Its protocol was fixed before any segmentation outer-fold score was observed, and every fold uses disjoint inner calibration for learning-rate, checkpoint, and decoder selection.
Across three pooled out-of-fold seed assemblages, BDL-Hard reaches $0.705\pm0.005$ mAP versus $0.586\pm0.011$ for cross-entropy segmentation and $0.286\pm0.008$ one-minute AP versus $0.071\pm0.002$.
The paired differences are 0.119 with a recording-bootstrap interval of $[0.108,0.130]$ for mAP and 0.215 with interval $[0.199,0.233]$ at one minute; BDL-Hard is higher on all five outer folds.
These intervals condition on the fitted models and fixed outer split, while the unanimous fold direction shows that the result is not confined to the earlier holdout partition.

\begin{table*}[t]
\centering
\caption{Frozen five-fold sleep comparison on identical outer and inner splits with three final seeds per fold. Values average pooled out-of-fold scores over seeds; 95\% intervals use 1,000 recording-level bootstrap replicates conditional on the fitted models and fixed outer split.}
\label{tab:grouped-sleep-segmentation}
\input{results/generated/grouped_sleep_segmentation_confirmation.tex}
\end{table*}

The separately calibrated holdout independently reproduces the direction and shows how it varies with tolerance (Figure~\ref{fig:sleep-main-results}).
Across three final seeds, BDL-Hard reaches $0.285\pm0.012$ one-minute AP versus $0.098\pm0.001$ for segmentation, or $2.9\times$ the segmentation AP.
The absolute difference is 0.186 with a series-bootstrap interval of $[0.150,0.225]$, conditional on the fitted models.
At three minutes the comparison is $0.577\pm0.006$ versus $0.349\pm0.011$; by thirty minutes the curves narrow to $0.836\pm0.002$ versus $0.817\pm0.013$ because both detectors often recover the episode when matching is permissive.
The tolerance curve therefore locates the main benefit at strict event detection, although AP still combines timing, confidence ranking, and duplicate suppression.

\subsection{The Boundary Formulation Accounts for Most of the Gain}

The held-out rerun gives BDL, boundary BCE, and segmentation the same learning-rate, checkpoint, and decoder search budget, selects each configuration on calibration data, and evaluates it once on a separate test partition.
Table~\ref{tab:confirmatory-sleep-holdout} shows test mAP of $0.694\pm0.004$ for BDL-Hard, $0.686\pm0.004$ for boundary BCE, and $0.593\pm0.009$ for segmentation over three fresh training seeds.
A paired series-cluster bootstrap gives a BDL-Hard minus boundary-BCE difference of 0.008 with a 95\% interval of $[0.003,0.013]$; the difference from segmentation is 0.101 with interval $[0.073,0.127]$.
Boundary BCE shares BDL's hard boundary channels, sparse initialization, and event decoder, and retains 92\% of the observed mAP gap over segmentation.
The large detector-level improvement is therefore associated with predicting boundaries directly; the Poisson score contributes a much smaller increment on this partition.
The nested study sharpens this decomposition: hard-boundary BCE reaches 0.702 mAP and retains 97\% of BDL-Hard's 0.119 gain over segmentation.

\begin{table*}[t]
\centering
\caption{Held-out sleep-GRU rerun with objective-specific calibration.
Values are test mAP and standard deviation over three fresh training seeds; the partition was not used for selection in this rerun.}
\label{tab:confirmatory-sleep-holdout}
\input{results/generated/confirmatory_sleep_holdout_compact.tex}
\end{table*}

The same five-fold design tests whether the smaller scoring-rule difference persists across partitions.
Each outer series fold uses disjoint inner calibration to select learning rate, checkpoint, and decoder before three final models are trained on all non-test series.
For hard targets, BDL reaches 0.705 pooled out-of-fold mAP and boundary BCE 0.702, a difference of 0.0034 with a 95\% interval of $[-0.0031,0.0112]$.
For unit-sum Gaussian targets, the comparison is 0.696 versus 0.687, a difference of 0.0089 with interval $[0.0004,0.0155]$.
Both comparisons favor BDL on three of five outer folds, but the Gaussian effect is concentrated in the final two; unweighted mean fold differences are 0.0019 and 0.0095, respectively.
Because pooled AP ranks outputs from independently calibrated fold models, its bootstrap interval does not replace variability across folds.
Taken together, the grouped results support at most a modest, target-dependent Poisson effect rather than a stable scoring-rule advantage.

\begin{table*}[t]
\centering
\caption{Five-fold sleep target-by-loss comparison. Each outer fold uses
independent calibration and three final seeds. Values are pooled out-of-fold
mAP; intervals use 1,000 series-cluster bootstrap replicates conditional on
the fitted models and fixed outer split.}
\label{tab:grouped-sleep-factorial}
\resizebox{0.72\textwidth}{!}{%
\input{results/generated/grouped_sleep_factorial.tex}
}
\end{table*}

\subsection{Alternative Explanations and Architecture Scope}

Class imbalance alone does not explain the development comparison.
With the GRU, split, scorer, and post-processing search fixed, BDL-Hard reaches 0.680 mAP, while cross-entropy, class-weighted cross-entropy, and focal segmentation reach 0.575, 0.544, and 0.525.
These controls reweight the interval mask but still recover event scores from state predictions after training.
BDL-Gaussian and BDL-Tolerance reach 0.660 and 0.605; these smoothed development runs predate the current crop-edge renormalization and remain provisional, whereas BDL-Hard is unaffected (Appendix~\ref{app:experimental-details}).

The offline bidirectional and convolutional development controls ask whether the detector comparison is tied to one backbone.
Direct boundary prediction is higher for the tested GRU, U-Net, and attention-gated U-Net under a shared event evaluator and decoder-search budget.
These are supporting development comparisons rather than nested architecture tests, and only the GRU receives the stronger held-out evaluation above.
The architecture ordering is not universal: a tested higher-capacity Transformer favors segmentation, 0.308 versus 0.013 mAP (Appendix Table~\ref{tab:transformer-strong-candidate}).
A protocol-frozen forward-GRU holdout also favors segmentation on average, $0.367\pm0.003$ versus $0.332\pm0.101$ for BDL-Gaussian.
A later right-sided-target study improves broad-tolerance mAP under a raw causal decoder but not under its predeclared trailing-filter comparison, and strict AP decreases.
These controls place the positive result in the tested offline recurrent and convolutional settings rather than in causal or self-attention models generally.

\subsection{Cross-Domain Scope}

CHB-MIT tests endpoint localization under a different signal modality and timing scale.
In five patient-grouped outer folds, BDL-Gaussian reaches 0.122 pooled out-of-fold mAP and matched Gaussian boundary BCE 0.108, but the 95\% patient-bootstrap interval for their difference is $[-0.0075,0.0348]$.
BDL is higher on only two folds, and the unweighted mean fold difference is $-0.0025$.
The patient-grouped study therefore establishes no stable scoring-rule ordering.

An exploratory recording-split, single-seed analysis shows that the boundary formulation can be competitive on high-rate EEG without supporting a patient-generalization claim.
At a two-second output stride, BDL-Gaussian reaches 0.360 mAP versus 0.252 for segmentation; at a one-second stride, BDL-Hard reaches 0.388 versus 0.287.
The preferred kernel changes with stride, and patient overlap plus the exclusion of seizure-free recordings prevent conclusions about unseen-patient screening.
Appendix~\ref{app:grouped-seizure-factorial} reports the grouped protocol, while Appendix~\ref{app:point-event-ablations} extends the construction to unpaired point events.

%% file: sections/result_figures.tex
\begin{figure*}[t]
\vspace{-0.8em}
\centering
\includegraphics[width=0.84\textwidth]{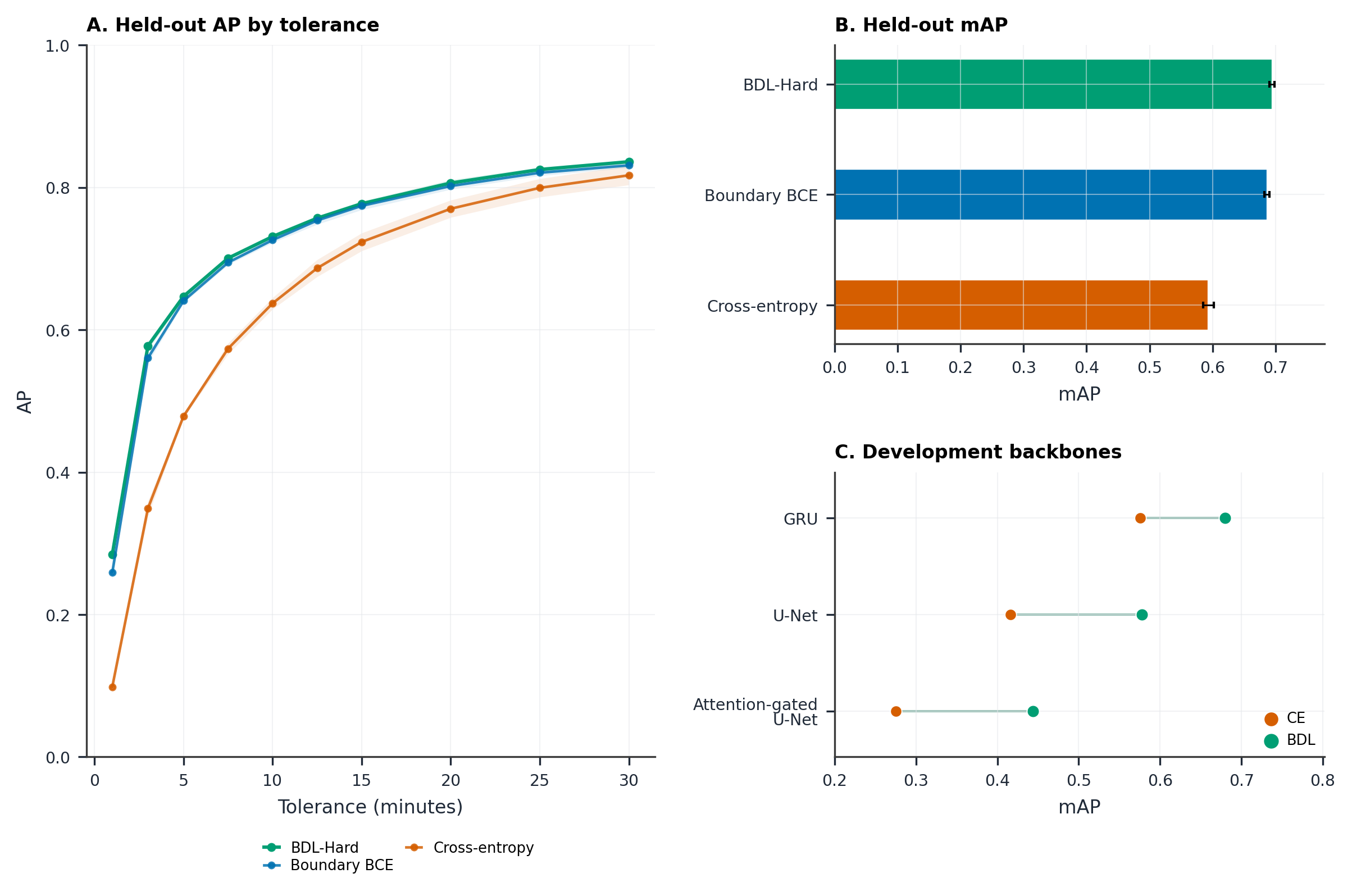}
\caption{Direct boundary detection improves strict-tolerance sleep-event AP.
Across three final seeds, BDL-Hard reaches $0.285\pm0.012$ one-minute AP versus $0.098\pm0.001$ for segmentation ($2.9\times$); the difference narrows at broad tolerances.
Panels A--B show held-out means and seed standard deviations. Panel C shows supporting development comparisons across offline bidirectional and convolutional backbones, not nested architecture tests.}
\label{fig:sleep-main-results}
\vspace{-0.8em}
\end{figure*}

%% file: results/generated/grouped_sleep_segmentation_confirmation.tex
\begin{tabular}{lcc}
\toprule
Method & OOF mAP & AP at 1 min \\
\midrule
BDL-Hard & 0.705 & 0.286 \\
Segmentation CE & 0.586 & 0.071 \\
\midrule
BDL $-$ segmentation & +0.119 $[0.108,0.130]$ & +0.215 $[0.199,0.233]$ \\
\bottomrule
\end{tabular}

%% file: results/generated/confirmatory_sleep_holdout_compact.tex
\begin{tabular}{lr}
\toprule
Method & Test mAP \\
\midrule
\textbf{BDL-Hard} & \textbf{0.694 $\pm$ 0.004} \\
Boundary BCE & 0.686 $\pm$ 0.004 \\
Segmentation CE & 0.593 $\pm$ 0.009 \\
\bottomrule
\end{tabular}

%% file: results/generated/grouped_sleep_factorial.tex
\begin{tabular}{lrrrr}
\toprule
Target & BDL & Boundary BCE & $\Delta$ & 95\% interval \\
\midrule
Hard & 0.705 & 0.702 & +0.0034 & $[-0.0031,0.0112]$ \\
Gaussian & 0.696 & 0.687 & +0.0089 & $[0.0004,0.0155]$ \\
\bottomrule
\end{tabular}

%% file: sections/limitations.tex
\section{Discussion}
\label{sec:limitations}

The results support a practical distinction between event detection and state estimation.
When evaluation consumes ranked onset, offset, or alarm times, directly predicting those events can be substantially better than learning interval occupancy and deriving transitions afterward.
When evaluation consumes samplewise occupancy, segmentation remains the direct objective.
This distinction, rather than the choice between Poisson and Bernoulli scoring, is the main empirical finding.

The boundary-versus-segmentation result is a detector-level comparison.
It changes labels, output channels, and decoding together, and weighted or focal losses do not exhaust the possible segmentation baselines.
The five-fold study reduces dependence on one partition, but its bootstrap intervals still condition on the fitted models and fixed outer split and do not include retraining or alternative split assignments; the separate holdout is corroborative rather than the sole basis for the claim.
The positive development direction across the bidirectional GRU and convolutional U-Nets is encouraging, but the Transformer reversal and protocol-frozen forward-GRU result show that architecture and optimization can dominate the choice of supervision.

The matched scoring-rule evidence is more narrowly identified.
Poisson and boundary BCE are nearly tied on hard targets and differ modestly on Gaussian targets across grouped sleep folds, with no stable ordering in patient-grouped CHB-MIT.
This proximity is expected in sparse bins: when a bin contains at most one event, its Bernoulli occurrence probability equals its mean hard count.
Poisson retains a clean additive interpretation and can represent multiple events per bin, but the present experiments do not show that this extra capacity improves AP.

Unit-sum kernels and sum binning conserve annotation mass in the target; they do not guarantee calibrated predicted counts.
The appendix count diagnostic finds a near-unity aggregate ratio but substantial series-level error, and a cross-fitted adapter does not improve it.
Likewise, BDL estimates a per-bin conditional mean rather than complete episode structure or a state-conditional transition hazard.
Peak selection, thresholds, endpoint pairing, and duration constraints remain decoder choices.

Finally, a conditional-mean score is not a proof of consistency for tolerance-matched AP.
Ranking, duplicate suppression, one-to-one matching, output stride, context, and optimization all affect the final detector.
The CHB-MIT study contains only seizure-positive recordings, and its recording splits contain patient overlap; the patient-grouped analysis is therefore the relevant scope check and does not establish screening performance.
The causal target-shape follow-up is similarly decoder-dependent: it improves broad-tolerance raw decoding but not the predeclared trailing-filter comparison or strict AP.
The current evidence supports retrospective offline localization in the reported sleep settings, while prospective alarms require causal models, latency-aware evaluation, and event-negative recordings.

%% file: sections/conclusion.tex
\section{Conclusion}

Event-time evaluation should be matched by event-time supervision.
BDL implements this principle with additive boundary targets, sum binning, and a per-bin Poisson conditional-mean score.
Across five nested sleep folds, BDL-Hard raises strict one-minute AP to $4.0\times$ the segmentation baseline and improves mAP by 11.9 percentage points, with positive differences on every outer fold.
A matched hard-boundary BCE detector retains 97\% of the mAP gap, while the same nested study shows only a modest, target-dependent difference between Poisson and Bernoulli scoring.
The principal result is therefore the value of predicting scored boundaries directly in the tested offline recurrent and convolutional settings; the scoring rule is a secondary choice, and causal, Transformer, and patient-grouped seizure results define the current limits.
More broadly, models should be trained to represent the sparse decisions that their evaluation and downstream users actually consume.

%% file: sections/appendix.tex
\section{Experimental Details}
\label{app:experimental-details}

Each reported comparison is specified by dataset, architecture, objective family, cross-validation split, and event-scoring configuration.
For each configuration, the selected hyperparameters, fold-level validation metrics, validation predictions, and validation-selected scoring settings are retained.
The summary tables and tolerance-specific AP curves are computed from those fold-level outputs.

\subsection{Training and Scoring Protocol}

Splits are made at the series level so windows from the same series are not split across train and validation.
Unless otherwise stated, sleep experiments use four folds, 20 training epochs, batch size 32, learning rate $3\times10^{-3}$, and validation every five epochs.
We separate optimization settings from decoder calibration.
Training hyperparameters are chosen at the benchmark-and-backbone level for stable multi-epoch learning and then held fixed inside matched objective comparisons.
Decoder parameters are selected on validation folds after training, because every objective must still be converted into ranked detections before event AP can be computed.
The decoder search includes score cutoffs, candidate smoothing, peak separation, and interval-alternation choices when they apply.
This makes each table a comparison of training signals under a shared validation protocol, not a collection of separately tuned detectors.
For sleep, three cutoffs are spaced uniformly from zero to the largest validation prediction for BDL and from zero to one for bounded outputs; candidate smoothing has Gaussian width zero or 10 original samples (zero or 50 seconds), peak separation is fixed at 360 original samples (30 minutes), and onset--wake-up alternation is either disabled or enabled.
The segmentation search also chooses between thresholded occupancy transitions and a centered difference-of-means transition score; this scorer family is selected on the same validation or calibration data.
For the reported CHB-MIT comparisons, decoder calibration uses expanded predictions on the original 256 Hz timeline.
The 11 cutoffs are defined analogously, candidate smoothing widths are 0, 256, and 512 samples (zero, one, and two seconds), peak separation is fixed at 145 seconds, and interval alternation is optional.
This keeps the decoder search aligned with second-scale seizure endpoint tolerances rather than reusing the broader sleep-scale smoothing grid.
When repeated seeds are available for the same configuration, tables report their mean and omit seed standard deviations for compactness.
The protocol is chosen to make objective comparisons controlled rather than to maximize any single configuration.
Four folds give each long recording a validation role while keeping full multi-epoch sweeps feasible.
Twenty epochs are long enough for the sleep validation curves to saturate in the optimization calibration, and validation every five epochs limits scoring overhead without changing the selected training objective.
The sleep learning rate comes from that Poisson-objective calibration; after selection, the value is shared with the matched segmentation controls.
We use one optimizer scale per benchmark and backbone rather than tuning a separate scale for every objective after seeing event AP.
This can leave some individual configurations below their best possible setting, but it prevents the objective comparison from becoming a post-hoc hyperparameter search.
Checkpoint selection follows the same rule for every configuration: validation event mAP is computed at the scheduled evaluation epochs, the best validation checkpoint supplies the saved predictions, and the same decoder search is then applied to those predictions.
The CHB-MIT high-capacity GRU configurations use a smaller learning rate, batch size 8, and gradient clipping because the EEG sequences are longer and the recurrent state is larger.
Batch size, clipping, and validation frequency are therefore memory, stability, and scoring-cost settings held fixed inside each benchmark comparison, not objective-specific advantages.
Output downsampling is selected by the metric-resolution and sparsity heuristic in Appendix~\ref{app:derivation}: the target has a consistent summed unit for any stride, but strict-localization claims require the output timeline to be fine enough for the matching tolerance.
When a coarser stride is used for memory or optimization, we treat it as a detector setting to be scored, not as a property guaranteed by the likelihood.

\begin{center}
\captionof{table}{Core training and scoring settings.
Settings are fixed within each matched benchmark comparison; CHB-MIT high-capacity configurations are rescored after training on the original 256 Hz timeline.}
\centering
\label{tab:core-protocol}
\resizebox{0.96\textwidth}{!}{%
\begin{tabular}{@{}lcccccc@{}}
\toprule
Benchmark setting & Model & Folds & Epochs & Batch & Learning rate & Output stride \\
\midrule
Sleep objective/backbone comparison & GRU / U-Net variants & 4 & 20 & 32 & $3\times10^{-3}$ & 50 s \\
CHB-MIT high-capacity & GRU 3x128 & 4 & 20 & 8 & $1\times10^{-3}$ & 2 s / 1 s \\
\bottomrule
\end{tabular}
}
\end{center}

Sleep samples are recorded every five seconds.  The model receives angle and
acceleration magnitude, hour-of-day and weekday categorical features, and
mean, maximum, minimum, and standard deviation summaries within each 50-second
output bin.  Training uses random seven-day windows.  Sleep preprocessing
follows the benchmark episode convention by retaining adjacent onset--wake-up
pairs; this omits 5 of 9,585 otherwise valid scored endpoints.  The BDL
formulation itself does not require paired endpoints, but the reported sleep
models use the retained 9,580 annotations.

CHB-MIT inputs are 21 bipolar EEG channels sampled at 256 Hz, with missing
values set to zero and amplitudes divided by 100.  Training uses random
one-hour windows, four summary statistics per output bin, and uniform
boundary jitter of at most 64 samples (0.25 seconds).  The evaluated subset
contains 141 seizure-positive recordings from 24 patients; seizure-free
recordings are absent, so the task is endpoint localization rather than
screening.
The experiments use version 3 (updated June 26, 2024) of the preprocessed
Kaggle snapshot
\url{https://www.kaggle.com/datasets/werus23/chb-mit-scalp-eeg-database-seizure-only},
not raw PhysioNet EDF files.  The 8,403,159,115-byte snapshot contains
\texttt{seizure\_events.csv} and 141 recordings under
\texttt{seizure\_256Hz\_dataset/}.  The CSV stores
\texttt{series\_id}, \texttt{onset}, and \texttt{offset}, with endpoint times
in seconds; each Feather file must contain the 21 channels enumerated in the
loader.  Its CSV SHA-256 is
\texttt{4d4aa3d99b9dab7774008d10ee566e87eebe96ef6b04667582d161ac0e2698a2}.
The SHA-256 of the sorted Feather checksum manifest is
\texttt{ba9a7cadf3114f8d8abe829807690b472f65162db8d096a9195e90a096e674a4};
the accompanying archive defines the canonical manifest encoding.

\begin{center}
\captionof{table}{Architecture specifications and trainable parameter counts.
Counts vary slightly with the one-channel segmentation or two-channel
boundary output; the table reports the boundary-output model.}
\centering
\label{tab:architecture-specs}
\resizebox{0.94\textwidth}{!}{%
\begin{tabular}{@{}llr@{}}
\toprule
Setting & Architecture & Parameters \\
\midrule
Sleep GRU & 2-layer bidirectional GRU, width 32 & 51,746 \\
Sleep U-Net & 3 levels, channels 64/128/256, kernel 7 & 560,258 \\
Sleep attention U-Net & same levels with skip attention & 37,150,658 \\
Sleep Transformer & 4 layers, width 64, 4 heads, dropout 0.1 & 201,602 \\
CHB-MIT GRU & 3-layer bidirectional GRU, width 128 & 1,201,410 \\
\bottomrule
\end{tabular}
}
\end{center}

All models use Adam with zero weight decay, a 10\% warmup from $10^{-6}$,
and cosine decay to $10^{-6}$.  Gradient norm is capped at 0.1 in the held-out
rerun and grouped studies.  The sleep Gaussian target uses $\sigma=50$ raw samples
(250 seconds), while CHB-MIT uses $\sigma=256$ samples (one second); both are
truncated at $3\sigma$ and normalized over visible support.  Focal
segmentation uses $\alpha=0.25$ and $\gamma=2$, and class-weighted
segmentation caps the positive-to-negative weight ratio at 100.  The held-out
and grouped sleep decoder grids use three score cutoffs, no or 50-second
smoothing, and optional onset--wake-up alternation.  CHB-MIT uses 11 cutoffs,
no, one-second, or two-second smoothing, and optional alternation.

BDL and matched boundary BCE use the same sparse output prior.  The implementation sets $\rate_0=\Delta/L$, where $\Delta$ is the output stride in raw samples and $L$ is a fixed benchmark reference spacing.  Sleep uses $L=17{,}280$ samples (one day), giving $\rate_0=5.79\times10^{-4}$ at stride 10; CHB-MIT uses $L=883{,}728$, giving $2.90\times10^{-4}$ and $5.79\times10^{-4}$ at strides 256 and 512.  These constants are fixed across folds rather than estimated separately on calibration or test partitions.  Table~\ref{tab:prior-ablation} shows that this initialization is operationally important.

\ifdefined\ANONYMOUSPAPER
An anonymized archive accompanying the submission contains the core implementation, deterministic split runners, unit tests, and key result snapshots.
\else
The implementation, deterministic split runners, unit tests, and result-collection code are available at \url{https://github.com/clarkipeng/EventDetectionPDF}; the workshop submission also carries a frozen anonymous code-and-result archive.
\fi

The current implementation preserves a trailing partial output bin and
renormalizes smoothed targets at recording edges.  All 15 saved run
configurations used lengths divisible by their output stride, so partial-bin
handling does not change a reported result.  Edge renormalization was added
after training the reported models.  A target audit found that it affects 5
of 9,580 retained sleep boundaries (0.1014 total missing target mass under the
earlier construction) and 1 of 396 CHB-MIT boundaries
($1.7\times10^{-5}$ missing mass).  The models were not retrained because the
affected mass is negligible, but those six annotations in the reported runs
did not satisfy exact recording-edge invariance.  This audit does not cover
random training-crop edges.  The reported smoothed sleep runs constructed
targets on the full recording and cropped them afterward, so kernels centered
near a crop edge could contribute less than one unit in that sampled window.
Current code constructs and renormalizes targets after cropping.  The
smoothed runs were not retrained; hard targets are unaffected.

At evaluation time, every trained model is converted to the ranked detection format required by the benchmarks.
BDL ranks local maxima of the learned score sequences.
Segmentation baselines derive event scores from changes in the predicted interval mask.
This conversion standardizes evaluation; BDL predicts the conditional mean
per-bin target, equal to a hard count or smoothed kernel-assigned mass.
The sleep benchmark uses minute-scale event matching windows from 1 to 30 minutes, while CHB-MIT uses second-scale matching windows from 1 to 60 seconds.

\section{Robustness and Scope Analyses}
\label{app:additional-results}

The appendix reports numerical values and controls for alternative explanations.
It first compares the boundary-detector formulation with interval segmentation, then uses matched boundary BCE to test the scoring rule more closely.

\input{sections/qualitative_figure}

\subsection{Direct Boundary Detection: Objective and Backbone Controls}

Tables~\ref{tab:sleep-target-comparison} and~\ref{tab:sleep-architecture-main} expand the main sleep result.
The first table isolates the training objective with the architecture fixed, and the second asks whether the BDL--segmentation gap persists when the backbone changes.
The objective sweep is deliberately narrow: all configurations share the same splits, encoder, event scorer, and validation-time decoder search.
The BDL configurations differ only in how each annotation is distributed before binning, while the segmentation configurations differ only in how the same interval mask is weighted during training.
This design holds model capacity and decoder-search budget fixed while comparing the two detector formulations.
The architecture sweep then repeats the most direct BDL and segmentation objectives across recurrent and convolutional sequence encoders.
The U-Net configurations test multi-scale convolutional representations that are natural controls for samplewise segmentation because they combine long context with time-localized decoding.
The attention-gated U-Net configuration remains within that convolutional family while adding learned selection over skip features before the final prediction.
Transformer diagnostics are reported separately because they change the detector family, sequence length, memory use, and optimizer tradeoffs at the same time.
For that reason, the self-attention results are reported separately under the same event evaluator and decoder-search budget rather than extending the main architecture table.

The objective table also includes two controls for alternative explanations of the main claim.
First, the BDL kernel configurations separate the Poisson objective from a generic smoothing trick: BDL-Hard uses an unsmoothed event at the annotated time, while the Gaussian and tolerance-aligned variants distribute the same one-event contribution over nearby bins.
The smoothed configurations check whether spreading target mass alone reproduces the hard-target result.
Second, the weighted and focal segmentation configurations keep the samplewise state labels and change only how errors on rare sleep-state samples are weighted.
If the main problem were only class imbalance, these segmentation controls should close the gap without changing the supervision from samplewise state labels to event occurrences.

\begin{center}
\captionof{table}{Sleep benchmark target comparison with a bidirectional GRU.
All objectives use the same splits and event-scoring protocol; mAP is the event-localization score used for model selection.}
\centering
\label{tab:sleep-target-comparison}
\resizebox{0.86\textwidth}{!}{%
\input{results/generated/sleep_objective_sweep.tex}
}
\end{center}

\begin{center}
\captionof{table}{Sleep benchmark architecture comparison.
Values are validation mAP aggregated over validation folds; $\Delta$ is BDL-Hard minus cross-entropy segmentation.}
\centering
\label{tab:sleep-architecture-main}
\input{results/generated/sleep_main_architecture.tex}
\end{center}

\IfFileExists{results/generated/attention_unet_control.tex}{%
\begin{center}
\captionof{table}{Attention-gated U-Net convolutional sensitivity check on the sleep benchmark.
Values are validation mAP aggregated over validation folds; $\Delta$ is BDL-Hard minus cross-entropy segmentation.}
\centering
\label{tab:attention-unet-control}
\input{results/generated/attention_unet_control.tex}
\end{center}
}{}

\clearpage
\subsection{Transfer, Smoothing, and Context Ablations}

Tables~\ref{tab:seizure-replication}--\ref{tab:online-ablation} organize additional evidence around transfer, smoothing choices, and available context.
Each table keeps the event-level scoring protocol fixed while changing one part of the detector setting.
The aim is not to claim a new best model for every domain, but to separate the contribution of the training target from the effects of signal type, architecture, decoder calibration, and causal information.
The studies therefore answer four questions: whether the same supervision transfers across signal domains, whether BDL is sensitive to sparse-output initialization or timing-uncertainty kernels, whether future context is needed to rank a boundary sharply, and whether outputs trained for event AP can be converted back to state sequences.

Table~\ref{tab:seizure-replication} is a compact transfer study from wrist accelerometry to CHB-MIT EEG while keeping the same objective families and ranked event-matching evaluation.
The seizure labels are still treated as event endpoints, but the signal, class rarity, sampling rate, and useful temporal scale differ substantially from sleep.
This transfer study deliberately moves the same objective family to sparse, high-rate EEG before changing capacity or output stride.
It therefore serves as a domain-shift control under the shared protocol.
Differences in this setting can reflect capacity, output stride, or decoder calibration as well as the training objective, so the later high-capacity and stride comparisons provide the main CHB-MIT evidence after those factors are controlled.

Tables~\ref{tab:prior-ablation} and~\ref{tab:target-width-ablation} isolate choices that are specific to BDL.
The prior ablation changes the initial output scale while leaving the one-event-per-annotation construction and Poisson loss unchanged.
The smoothing-width ablation changes only the support of the tolerance-aligned smoothing kernel before renormalization.
These comparisons test whether the main result depends on a particular rare-event initialization or a particular amount of label-time smoothing.
At full-recording construction, each annotation contributes one event in every configuration, so the intended amount of supervision is fixed across kernels; the crop-edge exception for reported smoothed runs is documented above.
The prior ablation is instead a calibration and optimization check: it asks whether the sparse Poisson output must be initialized near the event frequency for training to find useful peaks.
The width ablation asks a different question, namely how much timing uncertainty should be encoded around each annotation before training.
Together, these ablations identify whether performance is sensitive to output initialization or to the assumed label-time uncertainty, without changing the likelihood or the event occurrence assigned to each annotation.
Neither result changes BDL itself; these comparisons test practical calibration and smoothing choices around the same likelihood.

Table~\ref{tab:online-ablation} changes the information available to the encoder while keeping the event-occurrence construction and scorer fixed.
The offline GRU and U-Net configurations in the main text can use future samples inside the input window, whereas the forward recurrent models and causal Transformer only use current and past samples.
This study asks whether BDL remains effective when the representation has less temporal context for deciding an event time.
It isolates streaming context from the training signal: the loss can place event supervision on the scored output quantity, while the encoder may still lack evidence immediately after the transition that makes the boundary easy to rank.
The unsmoothed BDL streaming configurations underperform their segmentation controls: BDL-Hard reaches 0.219, 0.114, and 0.109 mAP for the forward GRU, forward LSTM, and causal Transformer, while the corresponding cross-entropy controls reach 0.288, 0.254, and 0.331.
This comparison does not identify whether the gap arises from context,
representation, optimization, decoding, or the objective.
A plausible mechanism is that a hard boundary target asks a causal model to
place a sharp spike before all confirming evidence is available, whereas
segmentation provides dense state supervision after the transition; the
experiments do not isolate that mechanism.
The smoothed BDL streaming analysis in Table~\ref{tab:online-smoothing-ablation} asks a narrower follow-up question: when future context is removed, does encoding label-time uncertainty help BDL recover sharper detections?
Smoothing improves the initial one-seed forward GRU and LSTM comparisons, with BDL-Gaussian reaching 0.298 and 0.232 mAP against 0.213 and 0.198 for cross-entropy, but it does not close the causal Transformer gap.
Tables~\ref{tab:online-smoothing-seed1-diagnostic}--\ref{tab:online-fgru-lr1e2-fairness} then test the forward-GRU case more directly.
At the post-hoc selected learning rate $10^{-2}$, BDL-Gaussian reaches 0.380 mean event mAP over three seeds, while a subsequently matched cross-entropy control reaches 0.342 with its validation-selected segmentation decoder.
The result establishes a viable forward-GRU configuration, not a general online advantage.
The Transformer controls continue to favor segmentation under their tested configurations.
The protocol-frozen forward-GRU follow-up in Table~\ref{tab:frozen-causal-sleep-holdout} resolves the post-hoc ambiguity against a general advantage.
Before any causal test score was computed, the existing split was pinned by hash; BDL-Gaussian, matched Gaussian boundary BCE, and segmentation received the common learning-rate grid $\{10^{-3},3\!\times\!10^{-3},10^{-2}\}$, checkpoint candidates at epochs 5, 10, 15, and 20, and the same calibration-only decoder search.
The selected configurations were retrained on the combined training and calibration partitions with seeds 11--13, then scored once on the 55 test series with valid event annotations.
Segmentation obtains the higher and much more stable mean, $0.367\pm0.003$ versus $0.332\pm0.101$ for BDL-Gaussian; the BDL-minus-segmentation difference is $-0.035$ with a 1,000-replicate paired series-bootstrap interval $[-0.065,-0.004]$.
That interval is conditional on the nine fitted models and does not integrate training-seed uncertainty: the paired seed differences are $+0.038$, $-0.154$, and $+0.010$.
The result therefore rejects a reliable causal BDL advantage under this protocol and identifies optimization stability as the main forward-GRU weakness.

A later study asks a narrower question within Gaussian BDL: whether moving the
training kernel's mass to the event time and future improves a forward GRU
relative to the original symmetric kernel.
The method was selected on a train-only development split after target-shape,
burn-in, optimization, pretraining, and count-supervision screens.
A three-fold check at 10 epochs did not satisfy its all-positive trailing-decoder
rule; a post-hoc 20-epoch run on the failed fold implicated the 40-update budget.
The final protocol therefore froze 20 epochs, three new seeds, and all decoder
modes before evaluating the 56-series test partition.
Table~\ref{tab:causal-target-final} retains the failed primary comparison:
trailing Gaussian decoding changes mAP from $0.358\pm0.030$ to
$0.347\pm0.009$, with paired differences $+0.015$, $-0.020$, and $-0.026$.
The predeclared parameter-free raw decoder instead improves every seed, from
$0.341\pm0.034$ to $0.379\pm0.013$, and centered smoothing reaches
$0.437\pm0.011$ versus $0.377\pm0.029$.
This gain is concentrated at broad tolerances: under raw decoding, one- and
three-minute AP decrease from 0.0029/0.0421 to 0.0019/0.0339.
Right-sided supervision also lowers count MAE from 7.03 to 6.30 events per
series-channel, but neither target construction guarantees calibrated counts.
The result is therefore evidence for target-shape sensitivity, not a general
strict-localization or online advantage.

\begin{center}
\captionof{table}{Compact CHB-MIT transfer study.
Values are validation mAP from the shared low-capacity configuration; Table~\ref{tab:seizure-highscore} gives the main stride-controlled CHB-MIT comparison.}
\centering
\label{tab:seizure-replication}
\input{results/generated/seizure_replication.tex}
\end{center}

\IfFileExists{results/generated/seizure_highscore_ready.tex}{%
The high-capacity CHB-MIT comparison separates two practical questions.
First, it tests whether the seizure benchmark needs a larger recurrent encoder than the compact GRU used in the shared transfer sweep.
Second, it tests whether output stride changes the objective comparison once the raw 256 Hz EEG timeline is compressed into a shorter model sequence.
This matters because endpoint matching is second-scale: a 256-sample stride gives a one-second model timeline at the original sampling rate, while a coarser stride can make optimization easier but changes localization quantization.
The fixed question is whether the same objective remains competitive when event duration, sampling rate, and evidence type change; the capacity and stride checks prevent that question from being confounded with a single low-capacity detector setting.
In both stride settings, at least one BDL variant exceeds cross-entropy under the relaxed CHB-MIT event-matching protocol.
The best variant also changes with stride: the Gaussian kernel is strongest at two seconds, while the hard target is strongest at one second.
This is evidence about detector sensitivity, not a universal kernel ranking: BDL remains competitive on high-rate EEG, while kernel choice, stride, and decoder calibration still matter.
\IfFileExists{results/generated/seizure_highscore_strict3_ready.tex}{%
The stricter one-to-three-second endpoint table sharpens the same conclusion.
At the one-second stride, BDL-Hard and BDL-Gaussian remain above cross-entropy
under strict endpoint matching, whereas BDL-Tolerance is essentially tied with
cross-entropy.
At the two-second stride, the Gaussian kernel is strongest under strict scoring.
This pattern is consistent with the stride heuristic: tight endpoint matching
rewards kernels and output timelines that keep peaks localized, while a
tolerance-shaped target can become too diffuse for the strictest boundary
windows.
}{}

\begin{center}
\captionof{table}{High-capacity CHB-MIT GRU stride ablation.
The two-second setting is the coarser compute--accuracy configuration; the one-second setting reduces boundary quantization under the same ranked event-scoring protocol.}
\centering
\label{tab:seizure-highscore}
\resizebox{0.86\textwidth}{!}{%
\input{results/generated/seizure_highscore_ready.tex}
}
\end{center}
\IfFileExists{results/generated/seizure_highscore_strict3_ready.tex}{%
\begin{center}
\captionof{table}{Strict CHB-MIT endpoint ablation.
Configurations use one-to-three-second boundary tolerances with the same high-capacity
GRU validation outputs as Table~\ref{tab:seizure-highscore}.}
\centering
\label{tab:seizure-highscore-strict}
\resizebox{0.86\textwidth}{!}{%
\input{results/generated/seizure_highscore_strict3_ready.tex}
}
\end{center}
}{}
}{}

\IfFileExists{results/generated/confirmatory_sleep_holdout.tex}{%
\subsection{Held-Out Decomposition of the Main Effect}
\label{app:confirmatory-holdout}

The main sleep results use matched cross-validation folds.  For this separate
rerun, we fixed a series-level train/calibration/test partition before fitting
the reported models: 165 series for training, 56 for calibration, and
56 for testing.  Each objective received the same learning-rate grid
$\{10^{-3},3\!\times\!10^{-3},10^{-2}\}$ and a single calibration seed.
The partition seed is 20260716, its train/calibration split uses 20260717, and model selection uses seed 101.
Calibration selected the checkpoint from epochs 5, 10, 15, and 20 and selected
the event decoder from the same cutoff, smoothing, and alternation grid used
throughout the sleep experiments.  We then retrained the selected
configuration on the combined 221 training and calibration series with three
seeds.  The test partition was used only for evaluation in this rerun.  One test
series has no valid scored boundary annotations, so the benchmark evaluator
scores the remaining 55 series.

Table~\ref{tab:confirmatory-sleep-holdout} adds two controls to BDL-Hard and
cross-entropy segmentation.  Boundary BCE predicts the same two unsmoothed
boundary channels as BDL-Hard but fits independent Bernoulli probabilities
with binary cross-entropy; it shares BDL's sparse output initialization and
event decoder.  This is the direct event-supervision control for the Poisson
score.  Localized MSE fits hard boundary maps with the repository's
earlier squared-error formulation.  We include MSE as a continuity and
mechanism ablation, not as a canonical time-series baseline.

BDL-Hard obtains $0.694\pm0.004$ test mAP, compared with
$0.686\pm0.004$ for boundary BCE, $0.593\pm0.009$ for segmentation, and
$0.579\pm0.013$ for localized MSE, where the deviations are across training
seeds.  A paired cluster bootstrap over the 55 scored test series (1,000
replicates, averaging the three seeds within each replicate) gives a BDL-Hard
minus boundary-BCE difference of $0.008$ with a 95\% interval of
$[0.003,0.013]$.  The corresponding difference from segmentation is $0.101$
with interval $[0.073,0.127]$.  These intervals are conditional on the three
fitted models and fixed partition; they do not include retraining or split
uncertainty.  The boundary-versus-segmentation comparison changes output
representation and decoding as well as labels, so it identifies a
detector-level effect rather than supervision alone.  Within that comparison,
the matched boundary formulation retains most of the observed improvement over
segmentation, while the Poisson score provides a smaller gain over an otherwise
matched boundary-BCE objective on this partition.

The selected configurations and their calibration scores are reported in
main-text Table~\ref{tab:confirmatory-sleep-holdout}.

To reduce dependence on that single partition, the grouped target-by-loss
study uses five outer series folds.  Within each outer training fold, a fresh
20\% series-level calibration partition selects learning rate from
$\{10^{-3},3\!\times\!10^{-3},10^{-2}\}$, checkpoint from epochs 5, 10, 15,
and 20, and the same event-decoder grid used by the held-out rerun.  The
selected configuration is then trained on all outer non-test series with
seeds 0--2.  Hard and unit-sum Gaussian targets are each crossed with the
Poisson and Bernoulli scores, so target smoothing is held fixed inside each
loss comparison.
The shuffled outer folds use seed 20260718; fold $j\in\{0,\ldots,4\}$ uses inner-split seed $20260719+j$ and selection seed $301+j$.
The soft boundary-BCE implementation constrains targets to $[0,1]$.
An audit of all 9,580 retained sleep boundaries found a maximum Gaussian target of 0.0079998 before binning and 0.079862 after sum binning, with no value above one; clipping therefore never activated, and the grouped Gaussian BDL and BCE runs used numerically identical targets.

The segmentation confirmation reuses the exact outer and inner splits,
learning-rate and checkpoint grids, final seeds, and decoder-search budget.
Its protocol was frozen before any segmentation outer-fold score was observed.
Across pooled out-of-fold predictions, BDL-Hard reaches 0.705 mAP versus 0.586
for segmentation and 0.286 versus 0.071 AP at one minute.  The paired
differences are 0.119 with a 95\% interval of $[0.108,0.130]$ and 0.215 with
interval $[0.199,0.233]$, respectively.  BDL-Hard is higher on all five outer
folds, whose mAP differences range from 0.096 to 0.128.

\begin{center}
\captionof{table}{Frozen five-fold BDL-Hard--segmentation comparison.  Values
average pooled out-of-fold scores over three seeds; 95\% intervals use 1,000
recording-level bootstrap replicates conditional on the fitted models and
fixed outer split.}
\centering
\label{tab:grouped-sleep-segmentation-appendix}
\input{results/generated/grouped_sleep_segmentation_confirmation.tex}
\end{center}

The final analysis pools out-of-fold detections across all five test folds and
averages the three seed-level mAP values for each objective.  Conditional
intervals use 1,000 bootstrap replicates over the 269 series with valid scored
boundaries; they hold the fitted models and outer split fixed.  Main-text
Tables~\ref{tab:grouped-sleep-segmentation} and
\ref{tab:grouped-sleep-factorial} report the detector and scoring-rule
comparisons; no outer test fold is used for configuration selection.

\begin{center}
\captionof{table}{Outer-fold BDL-minus-boundary-BCE mAP differences for the
grouped sleep comparison.  The final column is the unweighted fold mean.}
\centering
\label{tab:grouped-sleep-fold-differences}
\resizebox{0.98\textwidth}{!}{%
\input{results/generated/grouped_sleep_fold_differences.tex}
}
\end{center}

We also evaluated whether the additive output unit yields calibrated totals
on the grouped Gaussian outputs.  Before adaptation, summed rates have a
per-series-channel count MAE of 2.81 (relative MAE 0.439) and an aggregate predicted-to-observed
ratio of 1.005.  A regularized Poisson adapter trained with series-grouped
cross-fitting changes these values to 3.06 and 1.002, while pooled event mAP
changes only from 0.6956 to 0.6961.  This diagnostic distinguishes target
normalization from prediction calibration: BDL fixes the target unit, but the
reported detector does not accurately recover every recording-level count.
}{}

\IfFileExists{results/generated/confirmatory_seizure_holdout.tex}{%
As an exploratory scope check, we repeated the objective comparison in the strongest CHB-MIT detector
setting, the three-layer GRU with a one-second output stride, while using a
stricter patient-level partition.  Fourteen patients (81 recordings) are used
for training, five patients (29 recordings) for calibration, and five patients
(31 recordings) for testing.  BDL-Hard, BDL-Gaussian, boundary BCE, and
segmentation CE receive the same
learning-rate grid $\{3\!\times\!10^{-4},10^{-3},3\!\times\!10^{-3}\}$,
checkpoint schedule, and decoder grid.  The selected configurations are then
retrained on all 19 non-test patients with three seeds.
The partition and selection seeds are 20260716 and 101, respectively; final fits use seeds 0--2.
Because the main
CHB-MIT table uses recording-level validation folds, the absolute values in
Table~\ref{tab:confirmatory-seizure-holdout} are a separate patient-
generalization check and are not directly comparable with
Table~\ref{tab:seizure-highscore}.

All four objectives select learning rate $3\!\times\!10^{-3}$.
BDL-Gaussian and boundary BCE select epoch 15, while BDL-Hard and segmentation
select epoch 20.  Boundary BCE obtains $0.081\pm0.022$ test mAP,
BDL-Gaussian $0.078\pm0.017$, BDL-Hard $0.066\pm0.042$, and segmentation CE
$0.061\pm0.008$.
A paired patient-cluster bootstrap (1,000 replicates, averaging three seeds
within each replicate) gives a BDL-Gaussian minus boundary-BCE difference of
$-0.003$ with a 95\% interval of $[-0.016,0.039]$ and a BDL-Gaussian minus
segmentation difference of $0.018$ with interval $[-0.058,0.140]$.
For the hard target, BDL is below boundary BCE by $-0.0158$ with interval
$[-0.0560,-0.0001]$ on this partition.
With only five test patients, neither comparison establishes a patient-
generalization ordering.  Smoothing substantially reduces the BDL seed
variation and brings its point estimate close to boundary BCE, but the result
does not support a universal Poisson-score advantage.

\begin{center}
\captionof{table}{Exploratory CHB-MIT objective comparison on a patient-held-out
test partition.  Hyperparameters and the decoder are selected on calibration
patients; values are mean test mAP and standard deviation over three fresh
training seeds.}
\centering
\label{tab:confirmatory-seizure-holdout}
\resizebox{0.72\textwidth}{!}{%
\input{results/generated/confirmatory_seizure_holdout.tex}
}
\end{center}
}{}

\IfFileExists{results/generated/grouped_seizure_factorial.tex}{%
\subsection{Patient-Grouped CHB-MIT Transfer Check}
\label{app:grouped-seizure-factorial}

To reduce dependence on the five-patient held-out test partition, we repeated
the Gaussian target-by-loss comparison with five outer patient folds covering
all 24 CHB-MIT patients.  Each outer training set uses a disjoint 20\%
patient-level calibration split to select learning rate from
$\{3\!\times\!10^{-4},10^{-3},3\!\times\!10^{-3}\}$, checkpoint from epochs
5, 10, 15, and 20, and the decoder from the same no-, one-, and two-second
smoothing grid.  The selected BDL-Gaussian and matched Gaussian boundary-BCE
configurations use outer-split seed 20260718, inner-split seed $20260719+j$, and selection seed $301+j$ in fold $j$.
They are retrained on all outer non-test patients with seeds 0--2.
No test patient is used for model or decoder selection, and every patient
appears in exactly one outer test fold.

Pooling out-of-fold detections over 141 recordings gives BDL-Gaussian 0.122
mAP and boundary BCE 0.108.  The difference is 0.0138 with a 95\%
patient-cluster bootstrap interval of $[-0.0075,0.0348]$ from 1,000
replicates conditional on the fitted models and fixed patient split.  BDL is
higher on two of five outer folds, and the unweighted mean fold difference is
$-0.0025$.  Because pooled AP globally ranks outputs from independently
calibrated fold models, the positive pooled point estimate is not a robust
directional result.  The analysis does not establish a Poisson-score advantage
on CHB-MIT.

\begin{center}
\captionof{table}{Five-fold patient-grouped CHB-MIT Gaussian target-by-loss
comparison.  Each outer fold uses independent patient-level calibration and
three final seeds; the interval resamples 24 patients conditional on the
fitted models and fixed split.}
\centering
\label{tab:grouped-seizure-factorial}
\resizebox{0.82\textwidth}{!}{%
\input{results/generated/grouped_seizure_factorial.tex}
}
\end{center}
}{}

\IfFileExists{results/generated/decoder_sensitivity.tex}{%
\subsection{Decoder and Objective Sensitivity}
\label{app:postprocessing-likelihood-sensitivity}

Table~\ref{tab:decoder-sensitivity} asks whether the main event-AP
comparison depends on validation-time decoder tuning.
The default rows use raw model scores before validation-selected cutoffs,
smoothing, and alternation are applied; when segmentation has two default
boundary scorers, the better default scorer is used for the baseline.
The tuned rows use the validation-selected settings reported in the main
tables.
BDL remains above the matched segmentation baseline in both settings.
This check does not remove post-processing from the detector, because every
sequence output still needs a rule for producing ranked events.
It shows that the observed gap is not created only by the validation-time
decoder search.

\begin{center}
\captionof{table}{Decoder sensitivity for the main comparisons.
Default rows use pre-tuning scores; tuned rows use validation-selected
post-processing.}
\centering
\label{tab:decoder-sensitivity}
\resizebox{0.94\textwidth}{!}{%
\input{results/generated/decoder_sensitivity.tex}
}
\end{center}

\IfFileExists{results/generated/legacy_likelihood_ablation.tex}{%
Table~\ref{tab:legacy-likelihood-ablation} separates the unit-normalized
Poisson objective from earlier localized regression losses in the sleep GRU
objective sweep.
The legacy MSE objectives use localized hard, Gaussian, or tolerance-shaped
event maps with squared-error training; the BDL rows use the corresponding
unit-sum event construction with the Poisson score.
The hard-target comparison is the strongest evidence because it has three
paired seeds, and the Poisson objective is higher on each seed.
The Gaussian and tolerance rows are one-seed checks from the same collected
matrix, so they are included as directional diagnostics rather than as
standalone claims.

\begin{center}
\captionof{table}{Poisson score versus legacy localized MSE objectives on
the sleep GRU comparison.}
\centering
\label{tab:legacy-likelihood-ablation}
\resizebox{0.78\textwidth}{!}{%
\input{results/generated/legacy_likelihood_ablation.tex}
}
\end{center}
}{}
}{}

\begin{center}
\captionof{table}{Sparse-output-prior ablation on the sleep benchmark.
The sparse prior initializes the BDL output near the empirical rarity of the scored events; the target, loss, backbone, and decoder are unchanged.}
\centering
\label{tab:prior-ablation}
\input{results/generated/prior_ablation.tex}
\end{center}

\begin{center}
\captionof{table}{Tolerance-kernel width ablation on the sleep benchmark.
The multiplier rescales the tolerance-derived smoothing window before unit-sum normalization, testing label-time uncertainty without changing the total contribution of each annotation.}
\centering
\label{tab:target-width-ablation}
\input{results/generated/target_width_ablation.tex}
\end{center}

\subsection{Streaming and Architecture Controls}
\label{app:streaming-architecture-controls}

\begin{center}
\captionof{table}{Online-context ablation on the sleep benchmark.
Forward-only models remove future context, so these configurations measure how much the objective depends on future evidence rather than the central offline-localization setting.}
\centering
\label{tab:online-ablation}
\input{results/generated/online_ablation.tex}
\end{center}

\IfFileExists{results/generated/online_smoothing_ablation.tex}{%
\begin{center}
\captionof{table}{Smoothed-target streaming-context analysis on the sleep benchmark.
Configurations use the same forward-only and causal encoders as Table~\ref{tab:online-ablation}; $\Delta$ is the better smoothed BDL configuration minus cross-entropy segmentation.}
\centering
\label{tab:online-smoothing-ablation}
\input{results/generated/online_smoothing_ablation.tex}
\end{center}
}{}

\IfFileExists{results/generated/online_smoothing_seed1_diagnostic.tex}{%
Table~\ref{tab:online-smoothing-seed1-diagnostic} records a follow-up
single-seed forward-GRU learning-rate sweep prompted by the streaming-context
results.
With the same BDL-Gaussian target and decoder grid, increasing the learning
rate from $10^{-3}$ to $10^{-2}$ steadily improves validation mAP, default
event mAP, and tuned event mAP.
Performance improves with learning rate in this single-seed sweep, which is
consistent with underfitting at the lower learning rates but does not isolate
optimization from the other online-model differences.

\begin{center}
\captionof{table}{Forward-GRU online smoothing learning-rate diagnostic on the
sleep benchmark.
Rows use seed 1, BDL-Gaussian targets, 20 epochs, four validation folds,
and full decoder tuning over cutoff, smoothing, and interval alternation.}
\centering
\label{tab:online-smoothing-seed1-diagnostic}
\input{results/generated/online_smoothing_seed1_diagnostic.tex}
\end{center}
}{}

\IfFileExists{results/generated/online_smoothing_lr1e2_replication.tex}{%
Table~\ref{tab:online-smoothing-lr1e2-replication} repeats the selected
$10^{-2}$ forward-GRU setting across three seeds.
To avoid turning the replication into another decoder search, the score uses
the seed-1-selected operating point: cutoff $0$, Gaussian candidate smoothing
width $10$, and interval alternation.
The fixed decoder improves mean event mAP from 0.334 to 0.380, with seed
standard deviation 0.007 after smoothing.
The three-seed result is not attributable to one initialization, but the
learning rate and decoder were selected in the preceding post-hoc analysis.
We therefore treat it as an optimization sensitivity check rather than
confirmatory online evidence.

\begin{center}
\captionof{table}{Three-seed forward-GRU online smoothing replication on the
sleep benchmark.
Rows use BDL-Gaussian targets, learning rate $10^{-2}$, 20 epochs, four
validation folds, and the fixed seed-1-selected decoder setting
$(\mathrm{cutoff}=0,\ \mathrm{smooth}=10,\ \mathrm{alternating}=1)$.}
\centering
\label{tab:online-smoothing-lr1e2-replication}
\input{results/generated/online_smoothing_lr1e2_replication.tex}
\end{center}
}{}

\IfFileExists{results/generated/online_fgru_lr1e2_fairness.tex}{%
Table~\ref{tab:online-fgru-lr1e2-fairness} repeats the same learning-rate
change for the forward-GRU cross-entropy segmentation control.
The comparison uses the same seeds, folds, and training budget as the
BDL-Gaussian replication.
BDL is scored with the fixed seed-1-selected decoder from
Table~\ref{tab:online-smoothing-lr1e2-replication}, while cross-entropy keeps
its usual validation-selected segmentation decoder.
Under this comparison, BDL-Gaussian remains higher on every seed, with mean
event mAP 0.380 versus 0.342 for cross-entropy.
The result narrows the online interpretation: the improved forward-GRU BDL
result is not explained simply by giving only BDL the larger learning rate,
although the online setting remains a calibration-sensitive appendix analysis.

\begin{center}
\captionof{table}{Forward-GRU learning-rate fairness check on the sleep
benchmark.
Rows compare BDL-Gaussian and cross-entropy segmentation at learning rate
$10^{-2}$, using the same seeds, folds, 20 training epochs, and batch size 32.
BDL mAP uses the fixed decoder from
Table~\ref{tab:online-smoothing-lr1e2-replication}; CE mAP uses the
validation-selected segmentation decoder.}
\centering
\label{tab:online-fgru-lr1e2-fairness}
\input{results/generated/online_fgru_lr1e2_fairness.tex}
\end{center}
}{}

\IfFileExists{results/generated/frozen_causal_sleep_holdout.tex}{%
\begin{center}
\captionof{table}{Protocol-frozen forward-GRU sleep holdout.
Learning rate, checkpoint, smoothing, cutoff, and alternation are selected on
the calibration partition; test values are mean and standard deviation over
three fresh seeds.  The event evaluation contains 55 labeled test series.}
\centering
\label{tab:frozen-causal-sleep-holdout}
\input{results/generated/frozen_causal_sleep_holdout.tex}
\end{center}
}{}

\IfFileExists{results/generated/causal_target_final_test.tex}{%
\begin{center}
\captionof{table}{Final forward-GRU Gaussian-target shape study.
Values are test mean and standard deviation over three frozen seeds.
AP$_{1/3}$ reports mean one-/three-minute AP. Raw and trailing decoders are
causal; centered smoothing uses future lookahead. The trailing row is the
predeclared primary comparison and does not pass its promotion rule.}
\centering
\scriptsize
\label{tab:causal-target-final}
\input{results/generated/causal_target_final_test.tex}
\end{center}
}{}

\IfFileExists{results/generated/transformer_main_candidate_ready.tex}{%
A tuned offline-Transformer comparison tests whether the same supervision
effect appears under a self-attention encoder.
This configuration uses smoothed BDL targets and a lower learning rate chosen
for Transformer stability, while keeping the same sleep event-scoring protocol.
The comparison is interpreted separately from the convolutional controls
because it changes the detector family as well as the loss.

\begin{center}
\captionof{table}{Tuned offline-Transformer candidate on the sleep benchmark.
Values are validation mAP after validation-fold decoder calibration.}
\centering
\label{tab:transformer-main-candidate}
\input{results/generated/transformer_main_candidate_ready.tex}
\end{center}
}{}

\IfFileExists{results/generated/transformer_strong_candidate_ready.tex}{%
A higher-capacity offline Transformer provides a stronger self-attention
comparison for the sleep benchmark.
This six-layer, width-128, eight-head model uses batch size 8, a Transformer-
specific optimizer schedule, one seed, and all three BDL kernels.
This comparison asks whether BDL remains useful when the representation changes
substantially, without treating the Transformer as a drop-in replacement for a
U-Net skip-connection variant.
The current result is negative: all BDL variants are near 0.013 mAP, while
cross-entropy reaches 0.308.
Together with the smaller offline-Transformer diagnostic, this suggests that
the current self-attention detector is not a reliable vehicle for sparse
boundary-rate training without additional optimization, output-resolution, or
decoder work.

\begin{center}
\captionof{table}{Higher-capacity offline-Transformer candidate on the sleep benchmark.
Values are validation mAP after validation-fold decoder calibration; $\Delta$
is the best BDL configuration minus cross-entropy segmentation.}
\centering
\label{tab:transformer-strong-candidate}
\input{results/generated/transformer_strong_candidate_ready.tex}
\end{center}
}{}

\IfFileExists{results/generated/offline_transformer_diagnostic.tex}{%
This second table is not a duplicate of the higher-capacity candidate.
It uses a smaller four-layer, width-64, four-head model, batch size 32, three
seeds, and only BDL-Hard under the original shared training schedule.
It asks a narrower architecture question than the main recurrent and
convolutional controls and was not tuned as a new Transformer detector.
Under this shared setting, cross-entropy segmentation is substantially stronger
than BDL-Hard.
This marks a useful scope limit: BDL specifies how annotations become
per-bin event occurrences and how those predictions are scored, but architecture-specific optimization and calibration can still
be decisive when the representation is changed substantially.

\begin{center}
\captionof{table}{Offline-Transformer control on the sleep benchmark.
Configurations use the same event-scoring protocol as the main sleep experiments, but are
reported separately from the main recurrent and convolutional controls.}
\centering
\label{tab:offline-transformer-diagnostic}
\input{results/generated/offline_transformer_diagnostic.tex}
\end{center}
}{}

\IfFileExists{results/generated/local_conv_transformer_diagnostic.tex}{%
A follow-up causal Transformer diagnostic adds a local convolutional mixing
path to the self-attention block and scores the resulting BDL-Gaussian model
with the full validation decoder grid.
This test keeps the sleep event-scoring protocol fixed, but it changes the
architecture without a matched local-convolution segmentation control, so it
is reported only as a diagnostic.
Table~\ref{tab:local-conv-transformer-diagnostic} shows that the local-conv
Transformer reaches 0.344 mean tuned mAP over three seeds, compared with
0.303 before decoder tuning.
The higher BDL score shows that the tested Transformer result is sensitive to
architecture and tuning, but the unmatched diagnostic cannot recover an
objective comparison or explain the original failure.

\begin{center}
\captionof{table}{Local-conv causal Transformer diagnostic on the sleep benchmark.
Rows use BDL-Gaussian targets, 20 training epochs, four validation folds, and
the full decoder grid over cutoff, smoothing, and interval alternation.}
\centering
\label{tab:local-conv-transformer-diagnostic}
\input{results/generated/local_conv_transformer_diagnostic.tex}
\end{center}
}{}

\IfFileExists{results/generated/point_event_ablation_ready.tex}{%
\subsection{Point-Event Applicability}
\label{app:point-event-ablations}

The method does not depend on paired interval endpoints.
To check that the same event-occurrence construction applies to point-event data, we also run ablations on Martian bow-shock and credit-card fraud detection datasets from prior time-series event-detection work \citep{azib2023universal}.
These datasets score a single event type rather than onset--offset pairs, so interval alternation and duration constraints are absent.
Both checks use seed 0, four series-level folds, 20 epochs, and the same split for every objective.
Table~\ref{tab:point-event-ablation} reports the matched GRU comparison among BDL kernels and cross-entropy segmentation under the same ranked event-scoring protocol.
For these point-event configurations, the baseline is a samplewise event classifier: the positive label is placed at the event sample and peaks in the predicted event channel are ranked.
Thus the comparison removes interval masks and transition scoring and checks whether the implementation applies beyond paired boundaries.
These limited runs test an orthogonal structural question: whether the same score can train on labels with no interval state to segment.
Bowshock is nearly saturated for both objective families, while Fraud shows a larger BDL advantage.
These results do not imply that every point-event dataset favors BDL.
They illustrate that the construction can train on isolated scored events as well as interval boundaries; they do not establish effectiveness across point-event datasets.

\begin{center}
\captionof{table}{Point-event ablations.
Bowshock and Fraud contain a single scored event type, so they test BDL without paired interval endpoints or interval-validity post-processing.}
\centering
\label{tab:point-event-ablation}
\input{results/generated/point_event_ablation_ready.tex}
\end{center}
}{}

\subsection{State Reconstruction from BDL Outputs}
\label{app:state-reconstruction}

The sleep annotations also induce a binary sleep-state sequence, which lets us ask a different question from event AP: whether the learned outputs can reconstruct samplewise state labels.
This appendix analysis is separate from the main objective, because BDL is trained and selected for ranked onset and wake-up detection.
It is nevertheless useful because it checks whether BDL outputs can be reconciled with the segmentation view.
The conversion reverses the paper's main direction: the main method trains outputs for event scoring, whereas reconstruction asks how much of an interval mask can be recovered after training only on BDL outputs.
It is an analysis tool, and it does not require the number of sleep intervals at test time.
It is a controlled way to ask whether event-trained boundary channels contain enough transition information to recover a state sequence when a downstream user needs one.

For BDL we evaluate three conversions, each using only validation-selected post-processing parameters.
The peak-fill decoder first chooses ranked onset and wake-up peaks, then fills the state sequence between alternating onset--wake-up pairs.
This is closest to the event detector used for AP, because it starts from discrete scored detections and imposes interval legality afterward.
The integrated-output decoder instead works directly with the score sequences: it accumulates onset evidence minus wake-up evidence over time and thresholds the resulting cumulative sequence.
This tests whether the two boundary channels contain enough signed transition evidence to recover an interval mask without first committing to individual peaks.
The exponential score-to-transition mapping gives the smoothest conversion.
In a state model, a hazard would be the probability of a transition in the next output bin conditional on the current state.
The BDL outputs are conditional mean target values per output bin, not
state-conditional hazards, so this heuristic conversion introduces a scalar calibration before
mapping them to transition probabilities.
The onset output controls the probability of entering the sleep state, and the wake-up output controls the probability of leaving it,
\begin{align}
    p_t = p_{t-1}(1-\beta_t) + (1-p_{t-1})\alpha_t,
    \qquad
    \alpha_t = 1-\exp(-s\lambda^{\mathrm{on}}_t),
    \quad
    \beta_t = 1-\exp(-s\lambda^{\mathrm{off}}_t),
\end{align}
where $p_t$ is the induced sleep-state probability.
The scale $s$ is selected on validation folds and maps mean event-target scores into transition probabilities for this appendix diagnostic.
The recursion updates a state probability at every output bin, naturally allows zero, one, or multiple transitions over a long recording, and does not require knowing the number of sleep intervals in advance.
The initial state probability and final state threshold are also selected on validation folds.
Segmentation uses its direct state probabilities.
Table~\ref{tab:state-reconstruction} shows that BDL outputs can produce coherent state sequences, especially with the exponential recursion, but direct segmentation remains stronger on state-occupancy metrics.
This supports the paper's separation between event-localization supervision and samplewise state supervision.
BDL is better matched to ranked detections, while direct segmentation remains the preferred objective when evaluation consumes samplewise states.
The analysis changes the evaluation target after training and asks whether information learned for event localization can be integrated back into a state sequence.
It clarifies when direct segmentation remains preferred.

\begin{center}
\captionof{table}{State reconstruction analysis on the sleep benchmark.
Configurations use the same GRU validation outputs as the main objective comparison; F1 and IoU are computed on the induced binary sleep-state sequence.}
\centering
\label{tab:state-reconstruction}
\resizebox{0.94\textwidth}{!}{%
\input{results/generated/segmentation_reconstruction.tex}
}
\end{center}

\clearpage
\section{Proofs and Additional Derivations}
\label{app:derivation}

\subsection{Event Count Under Smoothing and Binning}

For the target in Equation~\ref{eq:smoothed-target},
\begin{align}
    \sum_{t=1}^{T} y_c(t)
    =
    \sum_{\tau\in B_c}\sum_{t=1}^{T}\kernel(t-\tau)
    =
    \sum_{\tau\in B_c}1
    =
    |B_c|.
\end{align}
The second equality uses the unit-sum kernel and finite-timeline renormalization.

\subsection{Output-Stride Heuristic}

The summation rule defines how to downsample the target, but it does not by
itself choose the output stride.
Let $\Delta$ be the number of original timesteps represented by one model
output bin and let $\delta_{\min}$ be the strictest matching tolerance used by
the event metric.
Because the decoded peak is piecewise constant at the bin scale, the event time
has a quantization uncertainty on the order of $\Delta/2$ after the prediction
is mapped back to the original timeline.
A conservative localization condition is therefore
\begin{align}
    \Delta \le \delta_{\min},
\end{align}
with a smaller stride preferred when strict-tolerance AP is the main metric.
This rule keeps the entire bin inside the smallest matching window.
A weaker quantization-only check is $\Delta/2\le\delta_{\min}$; it says a
centered peak can still fall within the strict tolerance, but it leaves little
margin for model error, smoothing, or decoder shifts.
This is a metric-resolution condition rather than a likelihood condition.
It is also an endpoint-localization condition, not an event-duration condition.
For CHB-MIT, the strict tolerance asks whether the reported onset or offset
time is close to the annotated endpoint; a seizure lasting many seconds can
still receive a poor strict endpoint score if its boundary is quantized or
shifted by the decoder.

There is also an event-sparsity condition.
For event type $c$, let $\rho_c$ denote the local event frequency measured
per original timestep.
The expected number of type-$c$ events in one output bin is
\begin{align}
    m_c = \rho_c\Delta.
\end{align}
Under a homogeneous-Poisson plug-in approximation,
\begin{align}
    \Pr[N_c(b)\ge 2]
    =
    1 - \exp(-m_c)(1+m_c)
    \approx
    0.5\,m_c^2,
    \qquad m_c \ll 1.
\end{align}
If one wants the probability of merging same-type events in one bin to be
at most $\eta$, this approximation gives the sufficient rule
\begin{align}
    \rho_c\Delta \le \sqrt{2\eta}.
\end{align}
Thus the output stride should keep $m_c$ well below one for each scored event
type whenever the goal is to rank individual event times.
The likelihood can still train on bins with counts larger than one, but those
bins merge multiple events into a single temporal cell and make
localization depend more heavily on the decoder.

Table~\ref{tab:stride-heuristic} applies these checks to the sleep
and high-capacity CHB-MIT settings.
In both sleep and CHB-MIT, the plug-in same-type collision probability is small
for the tested strides; this calculation is a heuristic, not a fitted event-process model.
The practical tradeoff is therefore not whether the Poisson objective can represent
the annotation count; it can.
The tradeoff is between metric resolution and the sequence length seen by the
encoder.

For sleep, the sampled timeline has a five-second step.
The main downsampling factor of 10 gives a 50-second model stride and about
25 seconds of half-bin quantization, which is below the strict one-minute
matching tolerance.
This setting is sparse and narrowly satisfies the conservative metric-resolution
condition.
For CHB-MIT, a factor of 512 gives a two-second stride at 256 Hz.
This is attractive computationally: one hour of EEG is reduced from 921600 raw
samples to 1800 model steps.
It also explains why the setting can perform well under relaxed seizure
matching.
However, its half-bin quantization is already one second, which equals the
strictest endpoint tolerance.
It satisfies the weaker centered-peak check but not the conservative full-bin
condition, so we treat it as a compute--accuracy configuration rather than as
the cleanest boundary-localization setting.

The matched high-capacity CHB-MIT factor of 256 gives a one-second model stride
and roughly 0.5 seconds of half-bin quantization.
It satisfies the conservative one-second rule and is therefore the better
setting for strict endpoint claims, but it doubles the recurrent sequence length
relative to the 512 setting.
We report the two high-capacity settings as a stride ablation because they test
different parts of the detector tradeoff: 512 asks whether the objective remains
useful in a lower-cost two-second-stride setting, while 256 asks whether the
same comparison holds when the output timeline is aligned to the strict
boundary metric.
Event duration does not replace this endpoint argument.
The annotated CHB-MIT seizures in our event file have minimum duration
6 seconds and median duration 45.5 seconds, so even a two-second model bin does
not collapse an entire seizure episode into one cell.
The endpoint metric is different: an onset or offset can still be late, early,
or poorly ranked relative to a one-second tolerance.
For boundary-localization claims we therefore treat one-to-three-second
CHB-MIT AP as the strict analysis and the wider five-to-sixty-second windows as
relaxed matching.

\begin{center}
\captionof{table}{Output-stride check for the sleep and CHB-MIT configurations.
$\Delta$ is the downsampling factor in original samples, strict tol. is the
smallest matching tolerance, $m$ is the expected same-type event count per bin
under the sparse-event prior, and $P_{\ge 2}$ is the small-$m$ collision
approximation.}
\centering
\label{tab:stride-heuristic}
\begin{tabular}{@{}lccccc@{}}
\toprule
Setting & $\Delta$ & Stride & Strict tol. & $m$ & $P_{\ge 2}$ \\
\midrule
Sleep benchmark & 10 & 50 s & 60 s & 5.79e-4 & 1.67e-7 \\
CHB-MIT high-capacity GRU & 512 & 2 s & 1 s & 5.79e-4 & 1.68e-7 \\
CHB-MIT high-capacity GRU & 256 & 1 s & 1 s & 2.90e-4 & 4.20e-8 \\
\bottomrule
\end{tabular}
\end{center}

\subsection{Conditional Mean Target}

For a fixed input, bin, and event type, write $m=\mathbb{E}[Y\mid\xseq]$.
The conditional expected BDL loss is
\begin{align}
    \mathbb{E}[\rate-Y\log\rate\mid\xseq]
    =
    \rate - m\log\rate + \mathrm{const}.
\end{align}
For $m>0$, the derivative is $1-m/\rate$ and the second derivative is $m/\rate^2>0$, so the unconstrained positive minimizer is $\rate=m$.
For $m=0$, the loss is increasing in $\rate$ and has infimum at $\rate\rightarrow0$.
Equation~\ref{eq:rate-param} imposes a numerical floor: within that output family the solution is clipped toward the lower bound, with an infimum at $\epsilon$ when $m\leq\epsilon$.

\subsection{Poisson Deviance Form}

For integer per-bin event counts, Equation~\ref{eq:bdl-loss} is the Poisson negative log-likelihood up to constants.
For smoothed fractional targets, we use the same parameter-dependent expression
as a Poisson deviance or quasi-likelihood score.
Replacing $Y!$ with $\Gamma(Y+1)$ gives a target-only continuous extension,
but not a normalized Poisson probability mass function over fractional values.
The omitted $\log\Gamma(Y+1)$ term therefore does not change optimization or
the conditional-mean optimum.

The loss also has a Bregman-divergence form.
Let $\phi(u)=u\log u-u$ for $u>0$, with the continuous extension $\phi(0)=0$.
Then
\begin{align}
    \rate - Y\log\rate
    =
    D_{\phi}(Y,\rate) - \phi(Y)
\end{align}
up to target-only constants, so minimizing the empirical BDL loss estimates
the conditional mean of the constructed target in each bin: a count for hard
targets and kernel-assigned event mass for smoothed targets.
Here $D_{\phi}(Y,\rate)=\rate-Y+Y\log(Y/\rate)$ is one-half of the Poisson unit deviance; summing it is a composite score unless conditional independence across bins is assumed.

%% file: sections/qualitative_figure.tex
\begin{figure*}[t]
\vspace{-0.6em}
\centering
\includegraphics[width=0.68\textwidth]{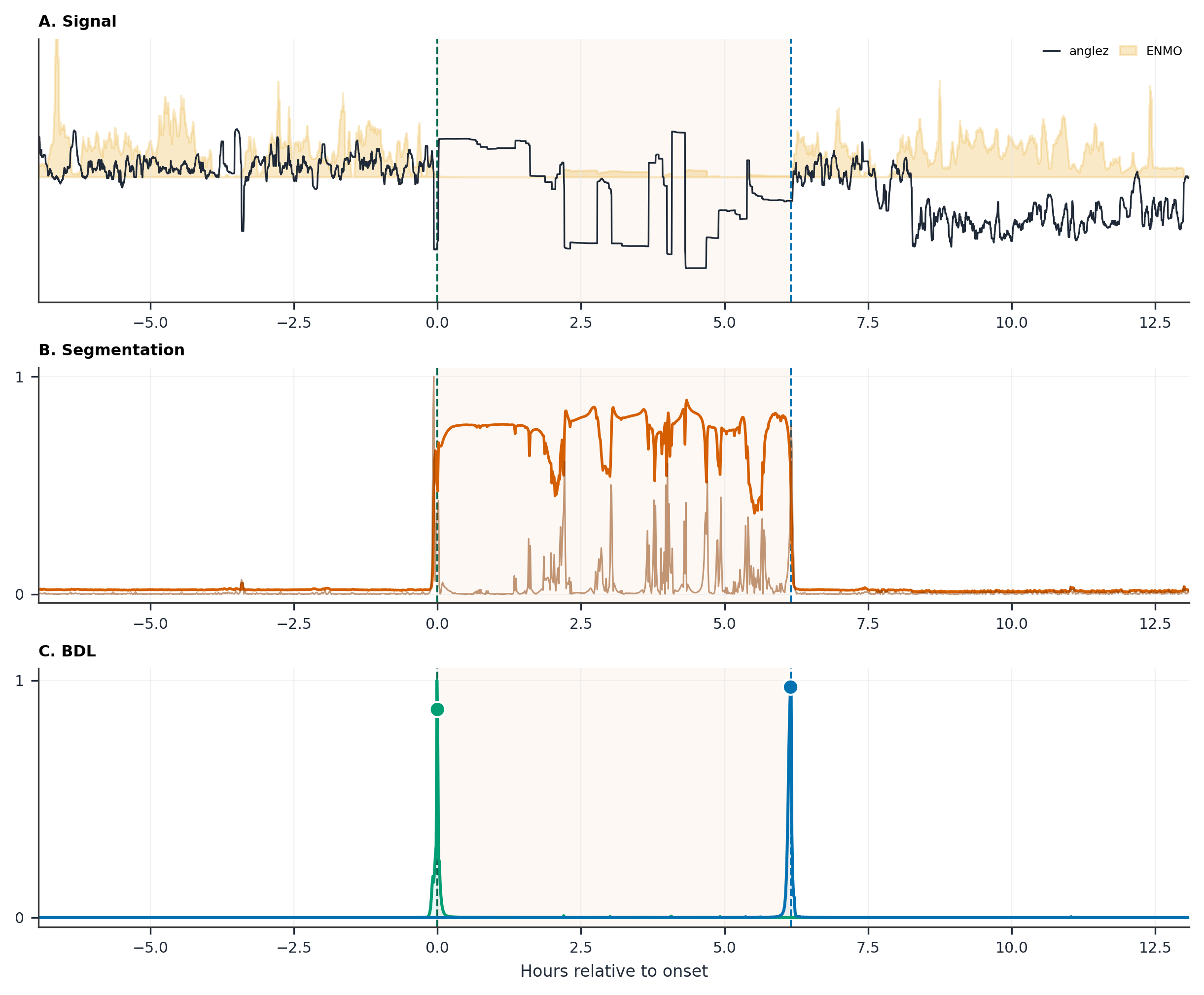}
\caption{Held-out sleep window.
The shaded span is the annotated sleep interval; dashed green and blue lines mark onset and wake-up.
Segmentation produces diffuse transition evidence after learning the interval state, while BDL-Hard produces localized endpoint peaks at the scored event times.}
\label{fig:qualitative-window}
\vspace{-0.8em}
\end{figure*}

%% file: results/generated/sleep_objective_sweep.tex
\begin{tabular}{lllr}
\toprule
Family & Method & Candidate decoder & mAP \\
\midrule
BDL & BDL-Hard & Event-score peaks & 0.680 \\
BDL & BDL-Gaussian & Event-score peaks & 0.660 \\
BDL & BDL-Tolerance & Event-score peaks & 0.605 \\
Segmentation & Cross-entropy & Segmentation transitions & 0.575 \\
Segmentation & Weighted cross-entropy & Weighted transitions & 0.544 \\
Segmentation & Focal loss & Focal transitions & 0.525 \\
\bottomrule
\end{tabular}

%% file: results/generated/sleep_main_architecture.tex
\begin{tabular}{llll}
\toprule
Model & BDL-Hard & Cross-entropy & $\Delta$ \\
\midrule
GRU & 0.680 & 0.575 & +0.104 \\
U-Net & 0.578 & 0.416 & +0.162 \\
Attention-gated U-Net & 0.444 & 0.275 & +0.168 \\
\bottomrule
\end{tabular}

%% file: results/generated/seizure_replication.tex
\begin{tabular}{llll}
\toprule
Model & BDL-Hard & BDL-Gaussian & Cross-entropy \\
\midrule
GRU & 0.010 & 0.018 & 0.031 \\
U-Net & 0.015 & 0.055 & 0.022 \\
\bottomrule
\end{tabular}

%% file: results/generated/seizure_highscore_ready.tex
\begin{tabular}{lllll}
\toprule
Setting & BDL-Hard & BDL-Gaussian & BDL-Tolerance & Cross-entropy \\
\midrule
GRU 3x128, 2 s stride & 0.317 & 0.360 & 0.267 & 0.252 \\
GRU 3x128, 1 s stride & 0.388 & 0.356 & 0.278 & 0.287 \\
\bottomrule
\end{tabular}

%% file: results/generated/seizure_highscore_strict3_ready.tex
\begin{tabular}{lllll}
\toprule
Setting & BDL-Hard & BDL-Gaussian & BDL-Tolerance & Cross-entropy \\
\midrule
GRU 3x128, 2 s stride & 0.120 & 0.138 & 0.048 & 0.036 \\
GRU 3x128, 1 s stride & 0.164 & 0.146 & 0.047 & 0.046 \\
\bottomrule
\end{tabular}

%% file: results/generated/grouped_sleep_fold_differences.tex
\begin{tabular}{lrrrrrr}
\toprule
Target & Fold 1 & Fold 2 & Fold 3 & Fold 4 & Fold 5 & Mean \\
\midrule
Hard & -0.0000 & +0.0115 & +0.0003 & +0.0006 & -0.0028 & +0.0019 \\
Gaussian & -0.0024 & -0.0004 & +0.0006 & +0.0321 & +0.0176 & +0.0095 \\
\bottomrule
\end{tabular}

%% file: results/generated/confirmatory_seizure_holdout.tex
\begin{tabular}{lrrrrr}
\toprule
Method & LR & Epoch & Cal. mAP & Test mAP & SD \\
\midrule
BDL-Hard & 3e-03 & 20 & 0.116 & 0.066 & 0.042 \\
BDL-Gaussian & 3e-03 & 15 & 0.169 & 0.078 & 0.017 \\
\textbf{Boundary BCE} & 3e-03 & 15 & 0.118 & \textbf{0.081} & 0.022 \\
Segmentation CE & 3e-03 & 20 & 0.069 & 0.061 & 0.008 \\
\bottomrule
\end{tabular}

%% file: results/generated/grouped_seizure_factorial.tex
\begin{tabular}{lrrrr}
\toprule
Benchmark & BDL-Gaussian & Boundary BCE & $\Delta$ & 95\% interval \\
\midrule
CHB-MIT & 0.122 & 0.108 & +0.0138 & $[-0.0075,0.0348]$ \\
\bottomrule
\end{tabular}

%% file: results/generated/decoder_sensitivity.tex
\begin{tabular}{lrrrrr}
\toprule
Comparison & Stage & Seeds & BDL & Baseline & $\Delta$ \\
\midrule
Sleep GRU & default & 3 & 0.674 & 0.512 & +0.162 \\
Sleep GRU & tuned & 3 & 0.680 & 0.575 & +0.104 \\
Sleep U-Net & default & 3 & 0.568 & 0.407 & +0.161 \\
Sleep U-Net & tuned & 3 & 0.578 & 0.416 & +0.162 \\
Sleep attention-gated U-Net & default & 3 & 0.438 & 0.248 & +0.190 \\
Sleep attention-gated U-Net & tuned & 3 & 0.444 & 0.275 & +0.168 \\
CHB-MIT 1 s stride & default & 1 & 0.367 & 0.198 & +0.169 \\
CHB-MIT 1 s stride & tuned & 1 & 0.388 & 0.287 & +0.101 \\
CHB-MIT 2 s stride & default & 1 & 0.341 & 0.215 & +0.127 \\
CHB-MIT 2 s stride & tuned & 1 & 0.360 & 0.252 & +0.109 \\
\bottomrule
\end{tabular}

%% file: results/generated/legacy_likelihood_ablation.tex
\begin{tabular}{lrrrrrr}
\toprule
Target & Seeds & Poisson & Legacy MSE & $\Delta$ & CI low & CI high \\
\midrule
Hard event target & 3 & 0.680 & 0.599 & +0.081 & +0.070 & +0.087 \\
Gaussian event target & 1 & 0.651 & 0.583 & +0.068 & -- & -- \\
Tolerance event target & 1 & 0.592 & 0.545 & +0.046 & -- & -- \\
\bottomrule
\end{tabular}

%% file: results/generated/prior_ablation.tex
\begin{tabular}{ll}
\toprule
Setting & mAP \\
\midrule
Sparse prior & 0.605 \\
No prior & 0.016 \\
\bottomrule
\end{tabular}

%% file: results/generated/target_width_ablation.tex
\begin{tabular}{ll}
\toprule
Target width & mAP \\
\midrule
$0.5\times$ & 0.640 \\
$1.0\times$ & 0.605 \\
$1.5\times$ & 0.563 \\
\bottomrule
\end{tabular}

%% file: results/generated/online_ablation.tex
\begin{tabular}{llll}
\toprule
Model & BDL-Hard & Cross-entropy & $\Delta$ \\
\midrule
Forward GRU & 0.219 & 0.288 & -0.069 \\
Forward LSTM & 0.114 & 0.254 & -0.140 \\
Causal Transformer & 0.109 & 0.331 & -0.222 \\
\bottomrule
\end{tabular}

%% file: results/generated/online_smoothing_ablation.tex
\begin{tabular}{lllll}
\toprule
Model & BDL-Gaussian & BDL-Tolerance & Cross-entropy & $\Delta$ \\
\midrule
Forward GRU & 0.298 & 0.258 & 0.213 & +0.086 \\
Forward LSTM & 0.232 & 0.153 & 0.198 & +0.034 \\
Causal Transformer & 0.170 & 0.182 & 0.233 & -0.051 \\
\bottomrule
\end{tabular}

%% file: results/generated/online_smoothing_seed1_diagnostic.tex
\begin{tabular}{lrrrr}
\toprule
Setting & Valid mAP & Default mAP & Tuned mAP & Tuned mF1 \\
\midrule
lr $=10^{-3}$ & 0.140 & 0.128 & 0.183 & 0.296 \\
lr $=3{\times}10^{-3}$ & 0.287 & 0.278 & 0.321 & 0.485 \\
lr $=5{\times}10^{-3}$ & 0.315 & 0.306 & 0.345 & 0.498 \\
lr $=7{\times}10^{-3}$ & 0.334 & 0.331 & 0.375 & 0.525 \\
lr $=10^{-2}$ & 0.345 & 0.338 & 0.382 & 0.534 \\
\bottomrule
\end{tabular}

%% file: results/generated/online_smoothing_lr1e2_replication.tex
\begin{tabular}{lrrrr}
\toprule
Seed & Valid mAP & Default mAP & Fixed mAP & Fixed mF1 \\
\midrule
0 & 0.335 & 0.323 & 0.371 & 0.444 \\
1 & 0.345 & 0.338 & 0.382 & 0.457 \\
2 & 0.346 & 0.341 & 0.388 & 0.456 \\
\midrule
Mean & 0.342 & 0.334 & 0.380 & 0.452 \\
SD & 0.005 & 0.008 & 0.007 & 0.006 \\
\bottomrule
\end{tabular}

%% file: results/generated/online_fgru_lr1e2_fairness.tex
\begin{tabular}{lrrrrr}
\toprule
Seed & BDL valid & CE valid & BDL mAP & CE mAP & $\Delta$ \\
\midrule
0 & 0.335 & 0.223 & 0.371 & 0.331 & 0.040 \\
1 & 0.345 & 0.219 & 0.382 & 0.338 & 0.044 \\
2 & 0.346 & 0.228 & 0.388 & 0.358 & 0.031 \\
\midrule
Mean & 0.342 & 0.223 & 0.380 & 0.342 & 0.038 \\
SD & 0.005 & 0.004 & 0.007 & 0.011 & 0.006 \\
\bottomrule
\end{tabular}

%% file: results/generated/frozen_causal_sleep_holdout.tex
\begin{tabular}{lrrrrr}
\toprule
Method & LR & Epoch & Cal. mAP & Test mAP & SD \\
\midrule
BDL-Gaussian & 1e-02 & 20 & 0.351 & 0.332 & 0.101 \\
Boundary BCE-Gaussian & 1e-02 & 20 & 0.350 & 0.333 & 0.102 \\
\textbf{Segmentation CE} & 1e-03 & 10 & 0.446 & \textbf{0.367} & 0.003 \\
\bottomrule
\end{tabular}

%% file: results/generated/causal_target_final_test.tex
\begin{tabular}{@{}lccrcc@{}}
\toprule
Decoder & Symmetric mAP & Right-target mAP & $\Delta$ & Sym. AP$_{1/3}$ & Right AP$_{1/3}$ \\
\midrule
Raw (causal) & $0.341 \pm 0.034$ & $0.379 \pm 0.013$ & +0.038 & 0.0029 / 0.0421 & 0.0019 / 0.0339 \\
Trailing Gaussian (causal) & $0.358 \pm 0.030$ & $0.347 \pm 0.009$ & -0.011 & 0.0022 / 0.0228 & 0.0000 / 0.0004 \\
Centered Gaussian (lookahead) & $0.377 \pm 0.029$ & $0.437 \pm 0.011$ & +0.060 & 0.0018 / 0.0285 & 0.0027 / 0.0371 \\
\bottomrule
\end{tabular}

%% file: results/generated/transformer_strong_candidate_ready.tex
\begin{tabular}{llllll}
\toprule
Model & BDL-Hard & BDL-Gaussian & BDL-Tolerance & Cross-entropy & $\Delta$ \\
\midrule
Offline Transformer & 0.013 & 0.013 & 0.013 & 0.308 & -0.295 \\
\bottomrule
\end{tabular}

%% file: results/generated/offline_transformer_diagnostic.tex
\begin{tabular}{llr}
\toprule
Objective & mAP & Runs \\
\midrule
BDL-Hard & 0.031 & 3 \\
Cross-entropy & 0.391 & 3 \\
\bottomrule
\end{tabular}

%% file: results/generated/local_conv_transformer_diagnostic.tex
\begin{tabular}{lrrrr}
\toprule
Seed & Valid mAP & Default mAP & Tuned mAP & Tuned mF1 \\
\midrule
0 & 0.310 & 0.304 & 0.345 & 0.479 \\
1 & 0.308 & 0.305 & 0.347 & 0.489 \\
2 & 0.301 & 0.300 & 0.340 & 0.481 \\
\midrule
Mean & 0.306 & 0.303 & 0.344 & 0.483 \\
\bottomrule
\end{tabular}

%% file: results/generated/point_event_ablation_ready.tex
\begin{tabular}{llll}
\toprule
Dataset & Objective & mAP & F1 \\
\midrule
Bowshock & BDL-Hard & 0.988 & 0.937 \\
Bowshock & BDL-Gaussian & 0.988 & 0.939 \\
Bowshock & BDL-Tolerance & 0.983 & 0.935 \\
Bowshock & Cross-entropy & 0.987 & 0.939 \\
Fraud & BDL-Hard & 0.755 & 0.746 \\
Fraud & BDL-Gaussian & 0.734 & 0.746 \\
Fraud & BDL-Tolerance & 0.704 & 0.698 \\
Fraud & Cross-entropy & 0.532 & 0.489 \\
\bottomrule
\end{tabular}

%% file: results/generated/segmentation_reconstruction.tex
\begin{tabular}{lllllll}
\toprule
Method & Reconstruction & F1 & IoU & Brier & Precision & Recall \\
\midrule
BDL-Hard & Exponential score-to-transition recursion & 0.887 & 0.797 & 0.086 & 0.841 & 0.940 \\
BDL-Hard & Event-score peaks to intervals & 0.792 & 0.656 & 0.118 & 0.686 & 0.937 \\
BDL-Hard & Integrated event-score difference & 0.408 & 0.257 & 0.663 & 0.262 & 0.944 \\
BDL-Gaussian & Exponential score-to-transition recursion & 0.888 & 0.799 & 0.086 & 0.858 & 0.922 \\
BDL-Gaussian & Event-score peaks to intervals & 0.839 & 0.723 & 0.088 & 0.747 & 0.957 \\
BDL-Gaussian & Integrated event-score difference & 0.452 & 0.294 & 0.448 & 0.361 & 0.738 \\
BDL-Tolerance & Exponential score-to-transition recursion & 0.883 & 0.791 & 0.091 & 0.881 & 0.886 \\
BDL-Tolerance & Event-score peaks to intervals & 0.842 & 0.728 & 0.085 & 0.754 & 0.955 \\
BDL-Tolerance & Integrated event-score difference & 0.515 & 0.347 & 0.305 & 0.438 & 0.667 \\
Cross-entropy & Direct state probability & 0.914 & 0.841 & 0.035 & 0.874 & 0.957 \\
Weighted cross-entropy & Direct state probability & 0.911 & 0.836 & 0.039 & 0.865 & 0.962 \\
Focal loss & Direct state probability & 0.901 & 0.819 & 0.059 & 0.850 & 0.959 \\
\bottomrule
\end{tabular}